\documentclass[10pt]{article} 

\usepackage[preprint]{rlj} 

\usepackage{amssymb}            
\usepackage{mathtools}          
\usepackage{mathrsfs}           
\mathtoolsset{showonlyrefs}     
\usepackage{graphicx}           
\usepackage{subcaption}         
\usepackage[space]{grffile}     
\usepackage{url}                
\usepackage{lipsum}             

\usepackage{hyperref}
\usepackage{url}
\usepackage{pgfplots}
\pgfplotsset{compat=1.18}
\usepackage{wrapfig} 

\usepackage{siunitx}
\usepackage{pifont}
\usepackage{booktabs}
\usepackage{tabularx}

\usepackage[nogroupskip,acronyms,nopostdot,style=super,nonumberlist,toc]{glossaries}

\newacronym{ai}{AI}{artificial intelligence}%
\newacronym{drl}{DRL}{deep reinforcement learning}
\newacronym{dl}{DL}{deep learning}
\newacronym{ml}{ML}{machine learning}
\newacronym{rl}{RL}{reinforcement learning}
\newacronym{ad}{AD}{autonomous driving}
\newacronym{av}{AV}{autonomous vehicle}
\newacronym{dnn}{DNN}{deep neural network}
\newacronym{ann}{ANN}{artificial neural network}
\newacronym{nn}{NN}{neural network}
\newacronym{dqn}{DQN}{deep Q-network}
\newacronym{cnn}{CNN}{convolutional neural network}
\newacronym{rnn}{RNN}{recurrent neural network}
\newacronym{rdqn}{RDQN}{recurrent deep Q-network}
\newacronym{ddqn}{DDQN}{double deep Q-network}
\newacronym{marl}{MARL}{multi-agent reinforcement learning}
\newacronym{dmarl}{DMARL}{deep multi-agent reinforcement learning}
\newacronym{mdp}{MDP}{Markov decision process}
\newacronym{mlp}{MLP}{multilayer perceptron}
\newacronym{nlp}{NLP}{natural language processing}
\newacronym{cv}{CV}{computer vision}
\newacronym{ppg}{PPG}{phasic policy gradient}
\newacronym{vae}{VAE}{variational auto-encoder}
\newacronym{td}{TD}{temporal difference}
\newacronym{mal}{MAL}{multi-agent learning}
\newacronym{per}{PER}{prioritized experience replay}
\newacronym{a2c}{A2C}{advantage actor critic}
\newacronym{sg}{SG}{stochastic game}
\newacronym{mg}{MG}{Markov game}
\newacronym{pomdp}{POMDP}{partially observable Markov decision process}
\newacronym{pomg}{POMG}{partially observable Markov game}
\newacronym{dpomdp}{dec-POMDP}{decentralized partially observable Markov decision process}
\newacronym{nrmse}{NRMSE}{normalized root-mean-square error}
\newacronym{ppo}{PPO}{proximal policy optimization}
\newacronym{gae}{GAE}{generalized advantage estimate}
\newacronym{rpl}{RPL}{residual policy learning}
\newacronym{apf}{APF}{articial potential field}
\newacronym{lstm}{LSTM}{long short-term memory}
\newacronym{ftg}{FTG}{follow-the-gap}
\newacronym{il}{IL}{imitation learning}
\newacronym{rrt}{RRT}{rapidly-exploring random tree}
\newacronym{torcs}{TORCS}{The Open Racing Car Simulator}
\newacronym{sac}{SAC}{soft actor-critic}
\newacronym{fov}{FOV}{field-of-view}
\newacronym{de}{DE}{disparity extender}
\newacronym{dst}{DST}{dynamic sparse training}
\newacronym{iqm}{IQM}{interquantile mean}
\newacronym{bc}{BC}{behavior cloning}
\newacronym{softmoe}{SoftMoE}{soft mixture-of-experts}
\newacronym{gap}{GAP}{global average pooling}

\title{Im\textit{pool}a: The Power of Average Pooling for\\Image-Based Deep Reinforcement Learning}

\setrunningtitle{The Power of Average Pooling for Image-Based Deep Reinforcement Learning}

\author{Raphael Trumpp\textsuperscript{1}, Ansgar Sch{\"a}fftlein\textsuperscript{1}, Mirco Theile\textsuperscript{1}, Marco Caccamo\textsuperscript{1}}

\emails{\{raphael.trumpp,ansgar.schaefftlein,mirco.theile,mcaccamo\}@tum.de}

\affiliations{
$^{1}$\textbf{TUM School of Engineering and Design, Technical University of Munich, Germany}
}

\contribution{
This work proposes the Impoola-CNN as image encoder for image-based deep reinforcement learning (DRL).
Impoola-CNN is built upon the widely used Impala-CNN and enhances its network architecture by leveraging global average pooling (GAP).
}{
The state-of-the-art Impala-CNN image encoder does not utilize GAP.
In contrast, GAP is used in many popular network architectures in computer vision \citep{he2016deep, xie2017aggregated, huang2017densely, hu2018squeeze, liu2022convnet}.
}

\contribution{
Our analysis for the full Procgen Benchmark demonstrates that Impoola-CNN excels at generalization, especially in environments without agent-centered observations. 
We show that Impoola-CNN outperforms other works on scaled networks in DRL.
}{
The Procgen Benchmark \citep{cobbe2020leveraging} allows for the evaluation of generalization capabilities of image-based DRL agents, which is hard to assess in Atari games.
}

\contribution{
We identify reduced translation sensitivity in Impoola-CNN as a key distinction from Impala-CNN.
Moreover, we find that Impoola-CNN is affected by fewer dormant neurons. 
}{
GAP reduces translation sensitivity \citep{lin2014network} and is considered a strong inductive bias in computer vision. \citet{sokar2023dormant} identified the dormant neuron phenomenon, i.e., a large fraction of neurons yielding near-zero output during, as a cause of wide-ranging performance decrease in scaled networks in DRL.
}

\summary{
As image-based deep reinforcement learning tackles more challenging tasks, increasing model size has become an important factor in improving performance.
Recent studies achieved this by focusing on the parameter efficiency of scaled networks, typically using Impala-CNN, a 15-layer ResNet-inspired network, as the image encoder.
However, while Impala-CNN evidently outperforms older CNN architectures, potential advancements in network design for deep reinforcement learning-specific image encoders remain largely unexplored.
We find that replacing the flattening of output feature maps in Impala-CNN with global average pooling leads to a notable performance improvement. 
This approach outperforms larger and more complex models in the Procgen Benchmark, particularly in terms of generalization.
We call our proposed encoder model Im\textit{pool}a-CNN.
A decrease in the network's translation sensitivity may be central to this improvement, as we observe the most significant gains in games without agent-centered observations.
Our results demonstrate that network scaling is not just about increasing model size---efficient network design is also an essential factor.
}

\begin{document}

\maketitle  

\begin{abstract}
As image-based deep reinforcement learning tackles more challenging tasks, increasing model size has become an important factor in improving performance.
Recent studies achieved this by focusing on the parameter efficiency of scaled networks, typically using Impala-CNN, a 15-layer ResNet-inspired network, as the image encoder.
However, while Impala-CNN evidently outperforms older CNN architectures, potential advancements in network design for deep reinforcement learning-specific image encoders remain largely unexplored.
We find that replacing the flattening of output feature maps in Impala-CNN with global average pooling leads to a notable performance improvement. 
This approach outperforms larger and more complex models in the Procgen Benchmark, particularly in terms of generalization.
We call our proposed encoder model Im\textit{pool}a-CNN.
A decrease in the network's translation sensitivity may be central to this improvement, as we observe the most significant gains in games without agent-centered observations.
Our results demonstrate that network scaling is not just about increasing model size---efficient network design is also an essential factor.
We make our code available at \url{https://github.com/raphajaner/impoola}.
\end{abstract}

\section{Introduction}
\label{sec:introduction}
Recent works on \gls*{drl} have revealed that apart from algorithmic improvements, considerable performance increases can come from the network architecture and training approach of the used \glspl*{dnn}.
Notably, the Impala-CNN model \citep{espeholt2018impala}, a 15-layer \gls*{cnn} with residual connections, outperforms the previously widely used Nature-CNN\footnote{The \gls*{cnn} model used by \citet{mnih2015human}, which consists of three Conv2d layers with $\{32,64,64\}$ filters.} as image encoder for image-based \gls*{drl} substantially \citep{cobbe2020leveraging, schwarzer2023bigger, obando2024deep}. 
However, raising the parameter count of \glspl*{dnn} in \gls*{drl} does not obey scaling laws for better performance as found in other areas in \gls*{dl} \citep{kaplan2020scaling, zhai2022scaling}, e.g., scaling ResNets to up to 152 layers improves performance for image classification \citep{he2016deep}.

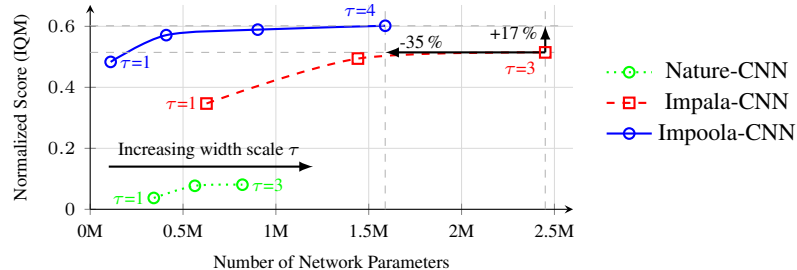
\begin{figure}[!t]
\hspace{1cm}
\centering
\begin{tikzpicture}
\begin{axis}[
    width=8cm, height=4.cm,
    xlabel={Number of Network Parameters},
    ylabel={Normalized Score (IQM)},
    xmin=0, xmax=2.600,
    ymin=0, ymax=0.67,
    xtick={0, 0.5, 1, 1.5, 2, 2.5},
    xticklabels={0M, 0.5M, 1M, 1.5M, 2M, 2.5M},
    ticklabel style = {font=\scriptsize},
    label style = {font=\scriptsize},
    grid=both,
    major grid style={line width=0.1mm, draw=gray!30},
    minor grid style={line width=0.1mm, draw=gray!20},
    axis lines=left,
    legend style={at={(1.05,0.4)}, anchor=mid west, draw=none, legend columns=1, font=\small}
]

\addplot[
    color=green, mark=o, thick, dotted, smooth, mark options={scale=1.0, solid, fill=blue!20}
] coordinates {
    (0.342448, 0.03723765903759162)
    (0.563712, 0.07709765958507639)
    (0.819792, 0.08073295660261579)
};
\addlegendentry{Nature-CNN}
\node[anchor=mid east, text=green] at (axis cs:0.342448, 0.03723765903759162) {\scriptsize $\tau$=1};
\node[anchor=mid west, text=green] at (axis cs:0.819792, 0.08073295660261579) {\scriptsize $\tau$=3};

\addplot[
    color=red, mark=square, thick, dashed, smooth, mark options={scale=1.0, solid, fill=red!20}
] coordinates {
    (0.626256, 0.34687364196416937)
    (1.441680, 0.494322883616673)
    (2.450640, 0.5145587055281597)
};
\addlegendentry{Impala-CNN}
\node[anchor=mid east, text=red] at (axis cs:0.626256, 0.34687364196416937) {\scriptsize $\tau$=1};
\node[anchor=north east, text=red] at (axis cs:2.450640, 0.5145587055281597) {\scriptsize $\tau$=3};

\addplot[
    color=blue, mark=o, thick, smooth, mark options={scale=1.0, solid, fill=blue!20}
] coordinates {
    (0.110160, 0.4833311527110104)
    (0.409488, 0.5711029978325005)
    (0.902352, 0.589188175098655)
    (1.588752, 0.6018528537775067)
};
\addlegendentry{Impoola-CNN}
\node[anchor=mid west, text=blue] at (axis cs:0.110160, 0.4833311527110104) {\scriptsize $\tau$=1};
\node[anchor=south east, text=blue] at (axis cs:1.588752, 0.5918528537775066) {\scriptsize $\tau$=4};

\draw[gray!58, dashed] (axis cs:0.0,0.6018528537775066) -- (axis cs:2.520,0.6018528537775066);
\draw[gray!58, dashed] (axis cs:0.0,0.5145587055281597) -- (axis cs:2.500,0.5145587055281597);

\draw[gray!58, dashed] (axis cs:1.588752,0.0) -- (axis cs:1.588752,0.65);
\draw[gray!58, dashed] (axis cs:2.450640,0.0) -- (axis cs:2.450640,0.65);

\draw[-latex, thick, black] (axis cs:2.450640,0.5145587055281597) -- (axis cs:2.450640,0.6018528537775066);
\node[anchor=east, text=black] at (axis cs:2.46,0.58) {\scriptsize +17\,\% };

\draw[latex-, thick, black] (axis cs:1.588752,0.5145587055281597) -- (axis cs:2.450640,0.5145587055281597);
\node[anchor=east, text=black] at (axis cs:1.95,0.55) {\scriptsize -35\,\% };

\draw[-latex, thick, black] (axis cs:0.100,0.14) -- (axis cs:1.200,0.14);
\node[anchor=south west, black] at (axis cs:0.100,0.135) {\scriptsize \text{Increasing width scale $\tau$}};

\end{axis}
\end{tikzpicture}
\caption{Impoola-CNN shows higher generalization over the full Procgen Benchmark than the Nature-CNN \citep{mnih2015human} and Impala-CNN \citep{espeholt2018impala} image encoders when scaling the network width $\tau$. By reducing the encoder's output dimension through GAP in Impoola-CNN, the number of weights in its Linear layers is reduced and, subsequently, the total parameters. The networks use a base configuration of $\{16,32,32\}$ filters per block. The presented results are obtained by evaluating the PPO agent on testing levels after 25M training steps. Normalized scores are aggregated as IQM across environments with 5 independent runs each, using different seeds.}
\label{fig:cover}
\end{figure}

There is high interest in finding methods for network scaling in image-based \gls*{drl} in recent studies \citep{nikishin2022primacy, schwarzer2023bigger,  sokar2023dormant, obando2024deep, ceron2024mixtures}.
Most of these works use the aforementioned Impala-CNN as the image encoder, typically scaling the network's width by increasing the output channels per Conv2d layer by a factor $\tau$.
Another line of work \citep{sinha2020d2rl, lee2024simba} has emphasized the impact of improved network design in particular to scale fully connected networks in regular \gls*{drl}.
Similar design improvements for image-based \gls*{drl} are of high practical appeal, primarily due to the prominence of end-to-end learning in robotic applications \citep{yang2021learning, funk2022learn2assemble, trumpp2023efficient}.

While experimenting with gradual magnitude pruning \citep{obando2024deep} for \gls*{drl}, we accidentally reduced only the Linear layer after the flattened feature maps in the Impala-CNN to a tiny fraction---the agent still performed well.
This finding motivated us to analyze the effect of the Flatten layer in more detail, analyzing the dormant neuron distribution in scaled Impala-CNN encoders.
\citet{sokar2023dormant} identified the dormant neuron phenomenon, i.e., a large fraction of neurons yielding near-zero output during \gls*{drl} training, as a potential cause hindering wide-ranging performance gains through network scaling.
After training a \gls*{ppo} \citep{schulman2017proximal} agent in the Procgen Benchmark \citep{cobbe2020leveraging} using the Impala-CNN, we found that the total amount of dormant neurons is especially prominent in the Linear layer \textit{after} the flattened features of the \gls*{cnn}-based encoder.
Notably, this layer also has a very high fraction of the network's overall weights placed due to the high-dimensional embedding created by the Flatten layer.
For image input of $64\times64$ pixels, the Impala-CNN has a total of 626,256 parameters, of which 83.76\,\% are in this Linear layer.

\textbf{The Power of Average Pooling:} 
We hypothesize that flattening the output feature maps in the \gls*{cnn}-based encoder is a root for training instabilities in image-based \gls*{drl} as it creates a high-dimensional embedding.
This hypothesis is in parallel to the results of \citet{sokar2024don} on \gls*{softmoe}, discovering that tokenizing the feature maps, which replace the Flatten layer, is key for the performance gain. 
In reference to this, we discover a significant architectural difference between the Impala-CNN and standard ResNet models \citep{he2016deep}:
There is a \gls*{gap} layer placed \textit{before} the block of Linear layers in standard ResNets.
This \gls*{gap} reduces the feature maps to single values, ensuring input size independence.
Moreover, this aggregates spatial information for reduced translation sensitivity while leading to a low-dimensional representation, potentially also enhancing gradient flow to earlier layers.
Motivated by these benefits, we propose extending the Impala-CNN with a GAP layer, naming the resulting model Im\textit{pool}a-CNN.
While this modification may seem \textit{subtle}, we show in Figure~\ref{fig:cover} that the Impoola-CNN outperforms the Impala-CNN for the Procgen Benchmark substantially while making efficient use of increased network widths.
Our results emphasize the value of \gls*{gap} in image-based \gls*{drl} with scaled networks.

Our main contributions are the following:
\begin{itemize}
    \item We identify architectural constraints in the Impala-CNN and propose the improved Im\textit{pool}a-CNN image encoder, which unlocks performance gains through efficient network scaling. 
    \item We provide extensive experiments for the full Procgen Benchmark with \gls*{ppo} and \gls*{dqn} agents. Our results show that our largest tested Impoola-CNN improves generalization in Procgen by 17\,\% while using 35\,\% fewer parameters than Impala-CNN.
    \item Our analysis investigates the effect of the \gls*{gap} layer on the network dynamics, identifying its decreased translation sensitivity as a characteristic quality of Impoola-CNN.
    \item The used code can be accessed at \url{https://github.com/raphajaner/impoola}.
    \end{itemize}
\section{Related Work}\label{sec:related_work}

\textbf{Network Scaling in Deep RL:}
For many control applications, the typically used fully connected networks do not or only marginally improve performance when scaling their parameter count \citep{henderson2018deep}.
However, recent works \citep{sinha2020d2rl, bjorck2021towards, nauman2024bigger} demonstrate that gains can be unlocked by improving the network architecture itself first before scaling, e.g., by introducing a residual block \citep{lee2024simba}.
Similarly in image-based \gls*{drl}, updating the standard Nature-CNN \citep{mnih2015human} encoder to the modern Impala-CNN model \citep{espeholt2018impala} yielded significant performance improvements for Atari \citep{schwarzer2023bigger, song2020observational} and Procgen games \citep{cobbe2019quantifying}.
\citet{song2020observational} compare different image encoder models in \gls*{drl}, 
highlighting the performance of Impala-CNN as models perform very differently in \gls*{drl} than supervised learning.
Further performance gains for the Impala-CNN model by scaling the network width are the subject of recent studies.
\citet{nikishin2022primacy, d2022sample, schwarzer2023bigger, sokar2023dormant} stabilize training through periodic reinitialization of the full network or neurons.
\citet{obando2024deep} show that performance for value-based \gls*{drl} is improved by unstructured gradual magnitude pruning during training.
\citet{ceron2024mixtures} propose replacing the encoder's Linear layer with a \gls*{softmoe} layer.
Further analysis by \citet{sokar2024don} identifies the tokenization of the feature maps for \glspl*{softmoe}, rather than the use of multiple experts, to drive the performance gain found in \cite{ceron2024mixtures}.

\textbf{Translation Invariance:}
A strong inductive bias in computer vision is to incorporate \textit{translation invariance}, i.e., invariance to the shifts of an object in the input image.
Intuitively, \gls*{gap} \citep{lin2014network} induces translation invariance, as the average operation is invariant to position \citep{mouton2020stride}.
However, Conv2d layers are not fully equivariant due to subsampling and boundary effects \citep{mouton2020stride}, e.g., they can exploit zero-padding and image orders to learn absolute positions \citep{islam2020much, kayhan2020translation}.
Thus, \glspl*{cnn} are not fully translation invariant even if \gls*{gap} is used. 
However, networks with \gls*{gap} are typically less sensitive to spatial translations of the input \citep{lin2014network}.
Translation sensitivity maps are a measure to quantify this property \citep{kauderer2017quantifying}.
\Gls*{gap} is still effective in practice and used in many popular network architectures \citep{he2016deep, xie2017aggregated, huang2017densely, hu2018squeeze, liu2022convnet}.

\textbf{Generalization in Deep Reinforcement Learning:}
Using the same environment for both training and testing results in high overfitting of \gls*{drl} agents \citep{zhang2018study, cobbe2019quantifying}.
Overfitting in \gls*{drl} may be associated with a loss of network plasticity \citep{nikishin2022primacy, sokar2023dormant} and generalization is theoretically closely related to invariances \citep{lyle2019analysis}.
The Procgen Benchmark \citep{cobbe2020leveraging} introduces various procedurally generated environments to quantify generalization. 
A number of invariance-based methods have been shown to facilitate generalization for Procgen environments, ranging from auxiliary losses \citep{raileanu2021decoupling} to data augmentation \citep{lee2020network, kostrikov2020image, raileanu2021automatic}.

\section{Background}\label{sec:background}

\subsection{Deep Reinforcement Learning}
The iterative optimization in model-free \gls*{drl} is formalized by a \gls*{mdp} with tuple $(\mathcal{S}, \mathcal{A}, \mathrm{T}, \mathrm{R}, \gamma)$.
Here, $\mathcal{S}$ and $\mathcal{A}$ represent the state and action spaces, respectively, while the transition function $\mathrm{T}: \mathcal{S} \times \mathcal{A} \rightarrow \mathcal{P}(\mathcal{S})$ defines the probability distribution over the next state given the current state and action.
The reward function is defined as $\mathrm{R}: \mathcal{S} \times \mathcal{A} \rightarrow \mathbb{R}$ and $\gamma$ is a discount factor. 
The mapping $\pi: \mathcal{S} \rightarrow \mathcal{P}(\mathcal{A})$ is called a stochastic action policy.
A \gls*{dnn} with weights $\theta$ parameterizes the policy $\pi_{\theta}$ in \gls*{drl}.
The optimal policy $\pi_{\theta}^{*}$ maximizes the expected return $V_{\pi_{\theta}}(s) = \mathbb{E}_{\pi_{\theta}}\left[\sum_{t=0}^{\infty} \gamma^t \mathrm{R}(s_t, a_t) \mid s_0=s\right]$.

\textbf{Q-Network Methods:}
These \gls*{drl} methods are typically based on an estimate of the Q-value function $Q_{\pi_{\theta}}(s, a):= \mathbb{E}_{\pi_{\theta}}\left[\sum_{t=0}^{\infty} \gamma^t \mathrm{R}(s_t, a_t) \mid s_0=s, a_0=a\right]$.
This function can be learned iteratively by temporal difference learning \citep{sutton1988learning} and bootstrapping the current Q-value estimate.
\gls*{dqn} \citep{mnih2015human} implements this by training a \gls*{dnn} with loss function $L(\theta) = \mathbb{E}_{(s, a, r, s') \sim \mathcal{D}} \left[ \left( r + \gamma \max_{a'} Q(s', a'; \bar\theta^-) - Q(s, a; \bar\theta) \right)^2 \right]$ where transitions $(s, a, r, s') \sim \mathcal{D}$ are sampled from the experience replay buffer $\mathcal{D}$ and by using a target network with $\theta^{-}$ as delayed copies of $\theta$.
Actions are obtained greedily by $a = \arg\max_a Q(s, a; \theta)$.

\textbf{Actor-Critic Methods:}
In addition to a critic network, e.g., $V(s; \phi)$ that estimates the state value, the action policy is defined as a dedicated actor network that can be directly optimized towards an optimization goal.
\Gls*{ppo} \citep{schulman2017proximal} is an \textit{on}-policy \gls*{drl} method, where the weights $\theta$ are updated with respect to the advantage function $A(s,a) = Q(s,a) - V(s)$. 
The \gls*{gae} \citep{schulman2015high} is the common choice to estimate $A(s,a)$.
The loss (clip version) of the \gls*{ppo} actor for a transition tuple $e=(s, a, r,s')$ of a trajectory $\tau=\{e, e', ...\}$ is given by $L(\theta) = \mathbb{E}_{\tau} \left[ \min \left\{ r(\theta) A, \text{clip}(r(\theta), 1 - \epsilon, 1 + \epsilon) A \right\} \right]$.
Here, $r(\theta) = \frac{\pi_\theta(a | s)}{\pi_{\theta_{\text{old}}}(a | s)}$ is the probability ratio between the old and new policy, where the hyperparameter $\epsilon$ limits their deviation.

\subsection{Convolutional Neural Networks}

\textbf{Impala-CNN:}
The Impala-CNN was introduced by \citet{espeholt2018impala} as a 15-layer model with residual connections specifically for encoding image inputs in \gls*{drl}.
The architecture combines two building blocks. 
As visualized in Figure~\ref{fig:impoola_architecture}, the ConvSequence $S_j$ blocks consist first of a Conv2d layer with MaxPooling and ReLU activation and then 2 subsequent ResBlock blocks as $S_j:\{C_{j} \xrightarrow{} P \xrightarrow{} R_{0,j} \xrightarrow{} R_{1,j}\}$; the ResBlock blocks $R_{i,j}$ are based on two Conv2d layers with ReLU activation and a residual connection.
The vanilla Impala-CNN stacks three ConvSequences $\{S_0, S_1, S_2\}$ with each block having the same amount of output channels $\{16,32,32\}$; scaled network versions multiply this configuration by a width scaling factor $\tau$.
The original implementation by \cite{espeholt2018impala} adds a Linear layer with 256 neurons to project the flattened feature map encodings $e$ to a fixed-dimension encoder output $z$ as part of the model.

\textbf{Pooling Layers:}
This fundamental operation in \glspl*{cnn} reduces the spatial dimensions of feature maps $\mathbf{x} \in \mathbb{R}^{C \times H \times W }$ where $C$, $H$, and $W$ represent channels, height, and width, respectively.
Average pooling computes the mean within a $k \times l$ window and stride $s$ as
\begin{equation}
\mathbf{y}(c, i, j) = \frac{1}{k\cdot l} \sum_{p=0}^{k-1} \sum_{q=0}^{l-1} \mathbf{x}\left(c, s \cdot i + p, s \cdot j + q\right),
\end{equation}
where $\mathbf{y}$ is the pooled output.
Typically, the window is square $k=l$, and the stride equals the window size.
\textit{Global} average pooling (GAP) \citep{lin2014network} reduces the spatial dimensions of feature maps $\mathbf{x}$ to a single value per map $\mathbf{y} \in \mathbb{R}^{C}$ by setting $k=H, l=W$.
\gls*{gap} reduces a network's translation sensitivity \citep{lin2014network} and simplifies its architecture by inducing no additional learnable parameters, facilitating scalable and efficient architectures.
Common machine learning frameworks provide adaptive implementations of Pooling layers where only the required output map dimension must be defined.
We refer to this by writing $\texttt{XPool(n,m)}$, which calculates the necessary $s,k,l$, and padding such that the output feature maps are of dimension $\mathbf{y} \in \mathbb{R}^{C\times n\times m}$.

\section{Impoola-CNN}\label{sec:methodology}
The Impoola-CNN is a \acrlong*{cnn} with an intended use as image encoder for image-based \gls*{drl}. 
Typically, the encoder output $z$ is fed into Linear layer prediction heads, e.g., an actor and critic head for \gls*{ppo}.
We discuss the essential design choices of this architecture below.

\begin{figure*}[!t]
    \centering
    \includegraphics[width=0.96\textwidth]{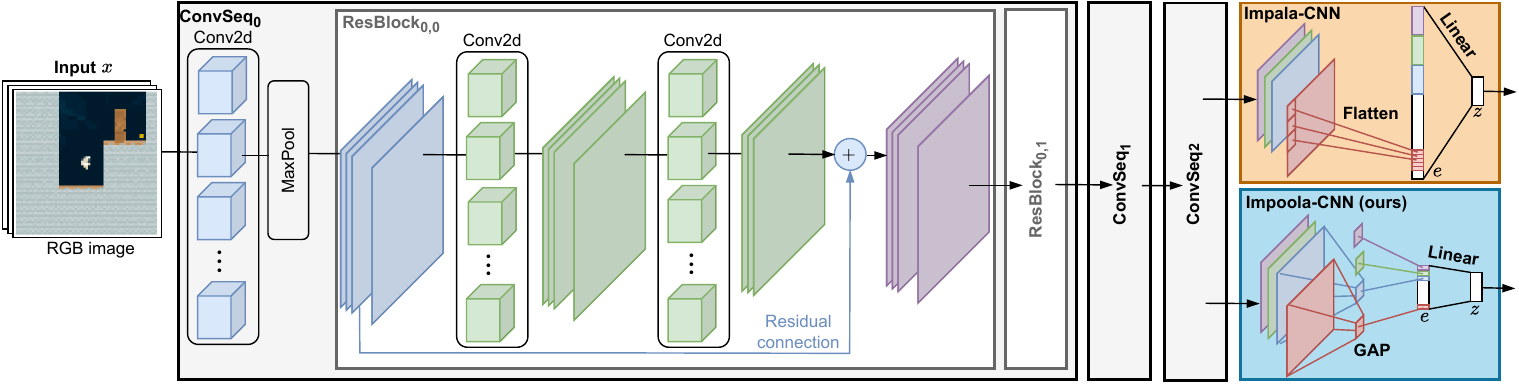}
    \caption{The Impala/Impoola-CNN models encode input images $x$ through a ResNet design, consisting of stacked ConvSeq blocks. ConvSeq blocks are built upon a first Conv2D layer, MaxPooling with stride 2, and two consecutive ResBlocks. ResBlocks are based on two Conv2d layers with a residual connection. The final feature maps are reduced to single values by using a GAP layer in Impoola-CNN, while the Impala-CNN flattens all features directly, resulting in a high-dimensional encoding~$e$. This encoding is then projected to the encoder's output variable~$z$ by a Linear layer.}
    \label{fig:impoola_architecture}
\end{figure*}

\textbf{Network Design:}
The Impoola-CNN builds upon the Impala-CNN from \citet{espeholt2018impala} and adds a \gls*{gap} layer after the last Conv2d layer as shown in Figure~\ref{fig:impoola_architecture}.
This change has a vast influence on the model architecture, as it reduces feature maps to single entries.
As listed in Table \ref{ap:tab:impoola_net}, the scaled Impala-CNN ($\tau=2$) consists of 1,441,680 learnable parameters for an input image of 64x64, of which 72.75~\% are located in the encoder's last Linear layer.
In contrast, for the same width scale~$\tau$, Impoola-CNN contains 409,488 parameters which are equally distributed over the network, with 4.06~\% in the Linear layer.
We speculate that this balanced distribution, specifically reducing the number of Linear layer weights, contributes to the performance increase of Impoola-CNN.

\textbf{Implementation Details:}
The Impala/Impoola-CNN encoders are deployed with an output feature dimension of $z \in \mathbb{R}^{256}$ in all experiments \citep{espeholt2018impala, huang2022cleanrl}. 
Setting the correct learning rate $\eta$ is crucial when comparing networks of different parameter counts. 
We searched a parameter space $\{7.5, 5.0, 3.5, 2.5, 1.0 \} {\scriptstyle\times 10^{-4}}$ for Impala/Impoola-CNN models, using a scaled version with $\tau=2$ as the base configuration.
We found that for \gls*{ppo}, both models work best with a learning rate of $\eta|_{\tau=2} = 3.5{\scriptstyle\times 10^{-4}}$, while setting $\eta|_{\tau=2}= 1.0{\scriptstyle\times 10^{-4}}$ is favorable for \gls*{dqn}.
We also tested this learning rate for scaled versions and concluded that we can obtain consistent results by adjusting the learning rate $\eta$ according to the following scaling rule $\eta|_{\tau}=\eta|_{\tau=2} \cdot {\frac{\tau}{2}}$, which was also shown to work well in \citep{obando2024deep}.

\section{Experiments}\label{sec:experiment}

\textbf{Experiment Design:}
We base our analysis on the Procgen Benchmark \citep{cobbe2020leveraging}. 
Our evaluation focuses on measuring the generalization of \gls*{drl} agents, for which Atari games are unsuitable.
Unless otherwise stated, the presented results are based on the \textit{full} benchmark with all 16 environments.
A qualitative description of the environment characteristics is given in Appendix \ref{ap:sec:procgen}.
The Procgen Benchmark allows for two tracks: \textit{efficiency}, where each level is sampled from the full level distribution, while the \textit{generalization} track restricts the levels per environment to a fixed set of 200 or 1000 levels for easy and hard settings, respectively.
We follow the training recommendation of \citet{cobbe2020leveraging} and train for 25M timesteps for the \textit{easy} and 100M for the \textit{hard} setting.

\textbf{Evaluation Metrics}:
We run periodic evaluations during training, gathering the episodic returns of 2,500 episodes.
We normalize episodic returns and report normalized scores $S$ using the normalization constants from 
\cite{cobbe2020leveraging} so that 1.0 corresponds to an optimal policy and 0.0 is equivalent to a random one.
For statistical relevance with a reasonable computational cost, we run all experiments for each environment with 5 independent runs using different seeds.
We presented results when aggregated across environments as \gls*{iqm} \citep{agarwal2021deep} scores and corresponding 95-\% stratified bootstrap confidence interval as shaded areas.

\textbf{Deep Reinforcement Learning Agents:}
This work uses \gls*{ppo} and \gls*{dqn} agents.
Our implementations are derived from CleanRL \citep{huang2022cleanrl} for PyTorch \citep{paszke2017automatic}.
The actor and critic for \gls*{ppo} share the image encoder.
Our \gls*{dqn} agent is extended by double Q-learning \citep{van2016deep}, multi-step rewards \citep{sutton1988learning}, and a simplified \gls*{per} \citep{schaul2015prioritized}.
Hyperparameters are listed in Appendix \ref{ap:sec:hyperparameters}.
Implementation details of other methods for benchmarking are given in Appendix \ref{ap:sec:benchmark_methods}.

\subsection{Evaluation}
Our initial experiment evaluates the effect of scaling the width of Impala-CNN and our proposed Impoola-CNN from $\tau=1$ to 3 and 4, respectively.
The results are shown in Figure~\ref{fig:cover} for \gls*{ppo} for the full Procgen Benchmark.
First, it can be seen that Impoola-CNN has substantially fewer total parameters at the same width levels $\tau$.
Overall, the largest tested Impoola-CNN achieves a 17\,\% higher IQM score for \textit{generalization} with 35\,\% less parameters than the Impala-CNN.
These results demonstrate the efficacy of the proportional weight distribution in the Impoola-CNN, as a high parameter count in the Linear layer does not directly translate into higher performance.

\begin{figure}
    \centering
    \includegraphics[width=0.8\textwidth]{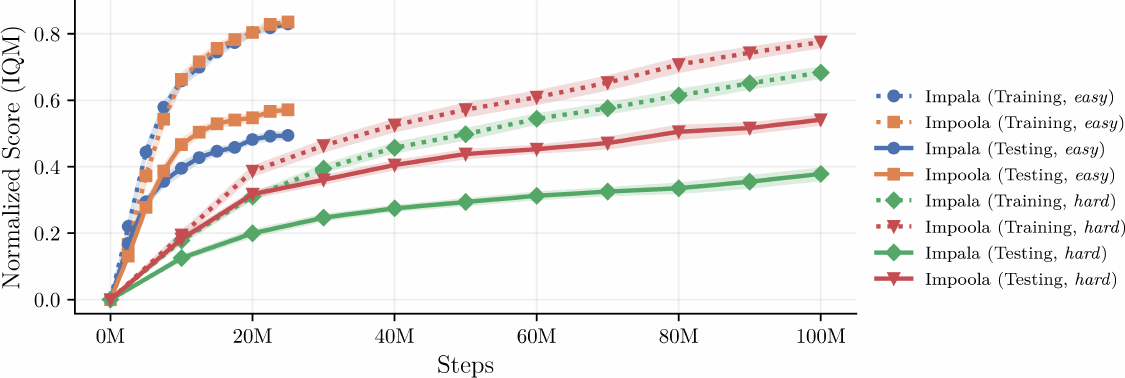}
    \caption{Generalization track for the \textit{easy} (blue and orange) and \textit{hard} setting (green and red) using PPO and scaled networks of $\tau=2$. Results for the levels used during training are depicted as dotted lines; test performance on unseen levels are solid lines. Agents in easy games are evaluated every 2.5M steps; the hard game setting requires longer training, so evaluations run every 10M steps. We utilize linear learning rate annealing in the hard setting to stabilize the long training duration.}
    \label{fig:generalization_easy_hard}
\end{figure}

Figure~\ref{fig:generalization_easy_hard} presents the results for the \textit{generalization} track with \gls*{ppo} in detail for $\tau=2$, comparing the scores for the \textit{easy} and \textit{hard} settings.
In addition to the usually only evaluated {testing} performance on the full distribution of levels, we display results for the restricted set of levels used in {training}.
In the \textit{easy} setting, the performance of Impala-CNN on \textit{training} levels trails the proposed Impoola-CNN only by a limited margin.
However, Impala-CNN's learned action policies generalize worse to unseen \textit{testing} levels than Impoola-CNN. 
This trend persists for the \textit{hard} setting as Impoola-CNN improves performance on unseen \textit{testing} levels but additionally outperforms Impala-CNN on the restricted \textit{training} levels. 
We posit that in the \textit{easy} setting, Impala-CNN is able to capture the fundamental game mechanics for high \textit{training} performance.
However, in contrast to Impoola-CNN, it fails to scale to the increased task complexity in \textit{hard}.
We hypothesize that its \gls*{gap} layer encourages the agent to learn more universal feature representations, which enable better adaptation to new levels and facilitate learning under challenging conditions.

\begin{figure*}[!t]
    \centering
        \begin{subfigure}{0.48\textwidth}
            \centering
            \includegraphics[width=0.8\textwidth]{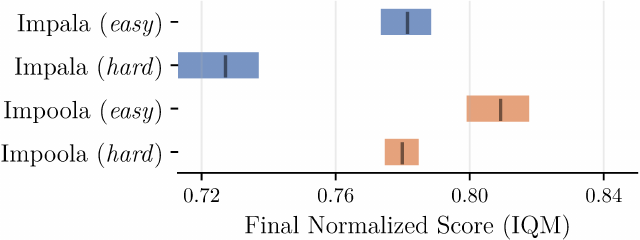}
            \caption{Efficiency tracks for PPO.}
            \label{fig:ppo_efficiency}
        \end{subfigure}
            \hfill
        \begin{subfigure}{0.48\textwidth}
            \centering
            \includegraphics[width=0.8\textwidth]{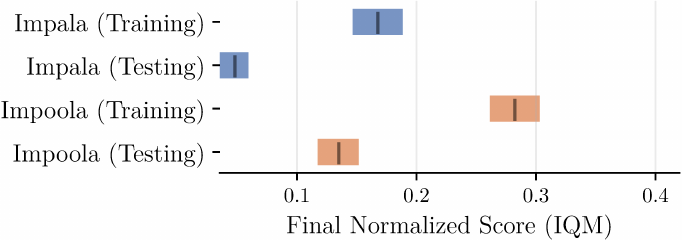}
            \caption{Generalization track (\textit{easy} setting) for DQN.}
            \label{fig:dqn_results}
        \end{subfigure}
    \caption{Further evaluation of Impala-CNN and Impoola-CNN encoders ($\tau=2$). We test \gls*{ppo} additionally for the efficiency track (\textbf{left}) and show results for \gls*{dqn}-based agents (\textbf{right}).}    
    \label{fig:further_results}
\end{figure*}

We conduct additional experiments, and first show results for \gls*{ppo} also in the \textit{efficiency} track, i.e., training on the full level distribution, in Figure~\ref{fig:ppo_efficiency}.
Impala-CNN's performance drops from the \textit{easy} to \textit{hard} setting meaningfully, while Impoola-CNN's remains more consistent.
This result is interesting since the \textit{efficiency} setting may favor larger networks due to their larger hypothesis space.
Nevertheless, the Impoola-CNN agent, consisting of 409,488 parameters, outperforms the Impala-CNN, which has approximately a 3x higher parameter count of 1,441,680.
Finally, Figure~\ref{fig:dqn_results} shows another experiment with \gls*{dqn} for the \textit{generalization} track (\textit{easy}).
While \gls*{dqn} achieves overall substantially lower performance than \gls*{ppo} in this track, the results affirm the previous trends.
Impoola-CNN not only improves testing performance for \gls*{dqn} agents but also training results.
An experiment with adding \gls*{gap} to the classical Nature-CNN can be found in the Appendix~\ref{app:add_experiments}, but with inconclusive results due to the Nature-CNN's overall weak performance and potential underparametrization.

\subsection{Benchmark}

\begin{figure*}[!t]
    \centering
    \begin{subfigure}{0.48\textwidth}
        \centering
        \includegraphics[width=0.8\textwidth]{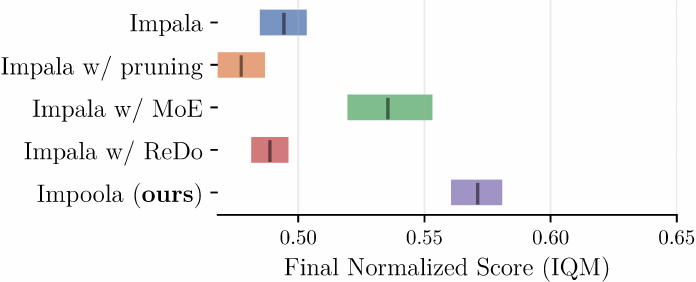}
        \caption{Benchmarking with other methods.}    
        \label{fig:benchmark}
    \end{subfigure}
    \hfill
    \begin{subfigure}{0.48\textwidth}
        \centering
        \includegraphics[width=0.8\textwidth]{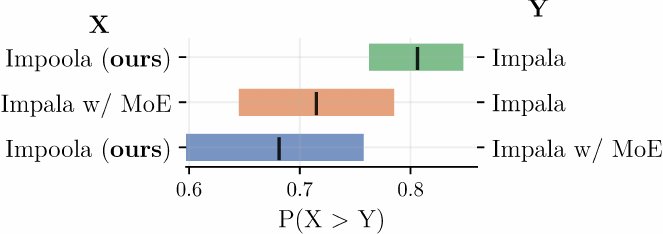}
        \caption{Probability of improvement that Algorithm X (left) performs better than Algorithm Y (right).}    
        \label{fig:testing_probability_of_improvement}
    \end{subfigure}
    \caption{Benchmarking the Impoola-CNN ($\tau=2$) against other methods for the generalization track (\textit{easy}) across the full Procgen Benchmark. We extend the Impala-CNN ($\tau=2$) by gradual magnitude pruning \citep{obando2024deep}, \gls*{softmoe} \citep{ceron2024mixtures}, and ReDo \citep{sokar2023dormant} for comparison. We present the final IQM scores (\textbf{left}) and the probability of improvement (\textbf{right}). The probability of improvement is a measure to estimate how likely it is that an algorithm outperforms another one in a single environment on average.
    }    
    \label{fig:benchmarking}
\end{figure*}

A comparison of Impoola-CNN to other recent methods related to network scaling is given in Figure~\ref{fig:benchmark} for the \textit{generalization} track (\textit{easy}).
It is evident that the gradual magnitude pruning method \citep{obando2024deep} and ReDo \citep{sokar2023dormant} do not translate to performance increases.
We presume that this situation is due to the fact that these methods were initially developed in the context of value-based \gls*{drl}, potentially magnified by our relatively short training period of 25M time steps in the easy setting.
However, using \gls*{softmoe} with 10 experts \citep{ceron2024mixtures} leads to a noteworthy increase compared to the vanilla Impala-CNN.
This observation is particularly significant, as \citet{sokar2024don} attribute the performance improvement of the method primarily to the tokenization of the encoder's output feature maps. 
This approach aligns conceptually to the \gls*{gap} layer in Impoola-CNN, as both mitigate the need for an excessively large Linear layer.
Impoola-CNN achieves the highest performance gains among the discussed methods and demonstrates the greatest likelihood of improvement over Impala-CNN in Figure~\ref{fig:testing_probability_of_improvement}.

\subsection{Understanding the Power of Average Pooling}
Introducing a \gls*{gap} layer in Impoola-CNN has two clear implications.
First, \gls*{gap} is well-known to reduce translation sensitivity in \glspl*{cnn} \citep{lin2014network}.
Second, the feature map encoding $e$ is reduced to the number of output feature maps, thus decreasing the connections to the subsequent fully connected Linear layer.
To understand these implications better, we evaluate related alternative approaches and extend the Impala-CNN instead of a \gls*{gap} layer, i.e., AvgPool(1,1), by AvgPool(2,2), MaxPool(1,1), or add a fourth ConvSeq block.
Moreover, we test a depthwise Conv2d layer, which creates $1{\times}1$ feature maps.

\begin{figure}[!t]
    \centering
    \includegraphics[width=\textwidth]{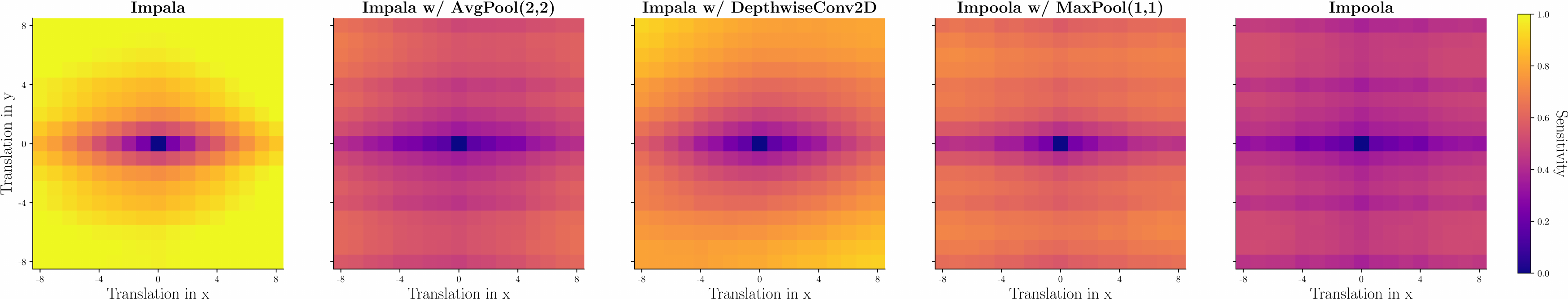}
    \caption{Translation sensitivity maps for the actor network of PPO in Bigfish (non-agent-centered game). The maps depict at each pixel (x,y) the corresponding sensitivity score that measures how the action probability distribution deviates when translating the input image by (x,y) pixels compared to the untranslated image. As the x and y axes are centered around 0, the center pixel's sensitivity score is always 0 as it references the untranslated image.
    Bright yellow colors indicate high translation sensitivity, i.e., the action probabilities differ substantially when translating the input by (x,y).}
    \label{fig:translation_sensitivity_maps}
\end{figure}

\begin{figure}[!t]
  \centering
  \begin{tikzpicture}
    \node[anchor=south west, inner sep=0] (grid) at (0,0) 
      {
        \includegraphics[width=0.9\textwidth]{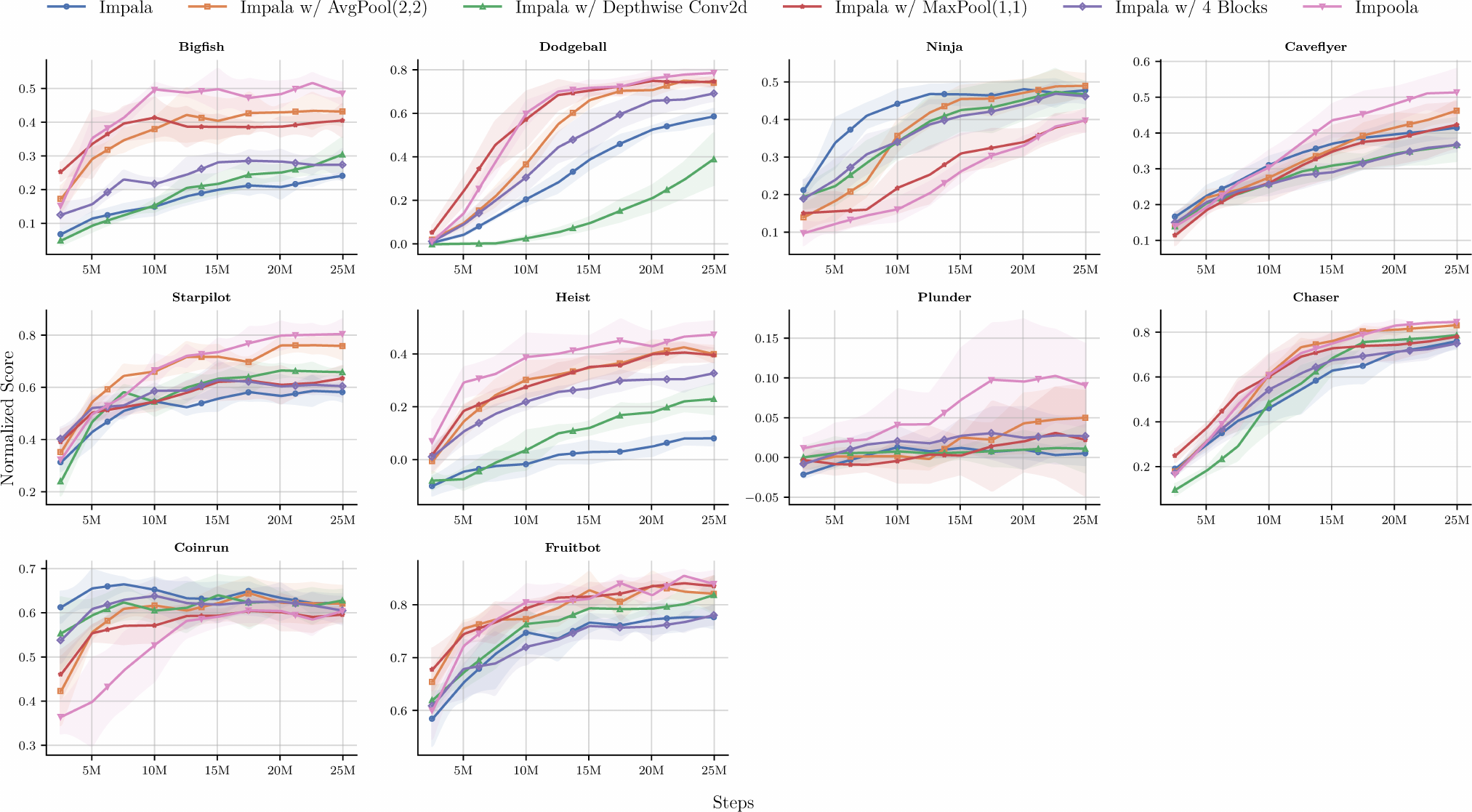}
      };
    
    \begin{scope}[x={(grid.south east)}, y={(grid.north west)}]
      \node[anchor=south west] at (0.52,0) 
        {
        \includegraphics[width=0.45\textwidth]{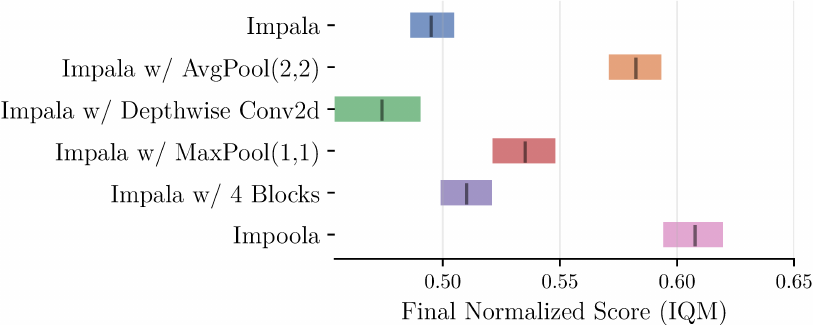}
        };
    \end{scope}
  \end{tikzpicture}
    \caption{Ablation study of Impoola-CNN ($\tau=2$) with results for generalization (\textit{easy}) using a subset of 10 Procgen games. Note that Caveflyer, Coinrun, and Ninja are environments with agent-centered observations. We show the final IQM scores (\textbf{bottom right}) and training curves with mean and standard deviation (\textbf{rest}).}
    \label{fig:ablation}
\end{figure}

\textbf{Translation Sensitivity:}
We quantify the translation sensitivity of the actor network, following a similar approach to \citet{kauderer2017quantifying}, by measuring the change in the action probability distribution when translating the input image by (x,y) pixels.\footnote{
 We measure translation sensitivity as the L1 distance between action probabilities from a Categorical distribution, computed from the actor’s output logits. This metric quantifies changes in action probabilities due to image translation and ensures comparability across networks via Softmax normalization.
 Only the agent and non-player characters are shifted against a stationary background to prevent artifacts.
 See Appendix~\ref{sec:translation_sensitivity_maps_appendix} for details.}
The corresponding translation maps are displayed in Figure~\ref{fig:translation_sensitivity_maps}.
Impala-CNN exhibits substantial variations in action distribution when the input image is shifted, leading to inconsistent actions.
In contrast, Impoola-CNN-based agents demonstrate reduced sensitivity to positional shifts, maintaining a stable action profile regardless of the absolute positions of entities in the image observation.
While MaxPooling(1,1) reduces the output feature maps to single entries like \gls*{gap}, the $\max$-operation appears to reduce the translation sensitivity less meaningfully.
We also see that AvgPool(2,2) is more sensitive than Impoola-CNN; the average pooling to (2,2) feature maps retains some spatial information.
Given that Depthwise Conv2D exhibits high translation sensitivity and performs substantially worse than models incorporating AvgPool(), despite identical parameter counts in the Linear layers, our hypothesis is further supported that translation insensitivity is a primary contributor to the observed performance gains.

\begin{figure*}[!t]
    \centering
    \begin{minipage}{0.58\textwidth}
        \centering
        \includegraphics[width=0.95\textwidth]{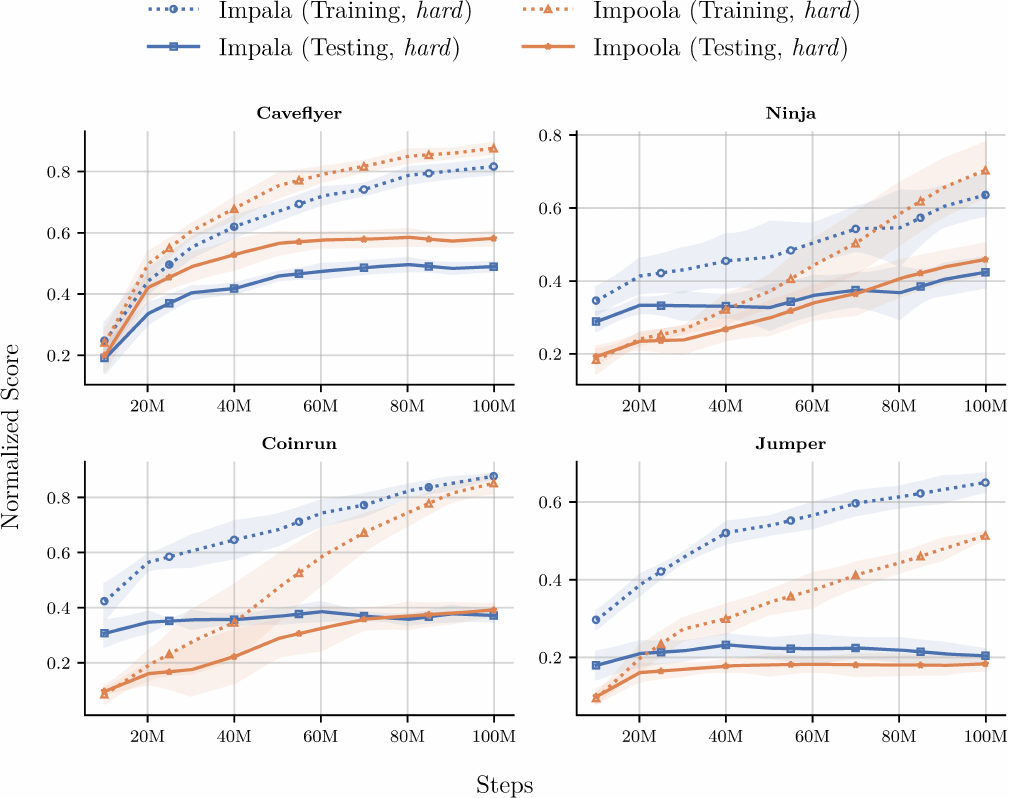}
        \caption{Results for games with agent-centered observations in the \textit{generalization} track (\textit{hard}) for PPO ($\tau=2$).}
        \label{fig:agent_centered_envs_generalization}
    \end{minipage}
    \hfill
    \begin{minipage}{0.38\textwidth}
    \centering
    \includegraphics[width=0.88\linewidth]{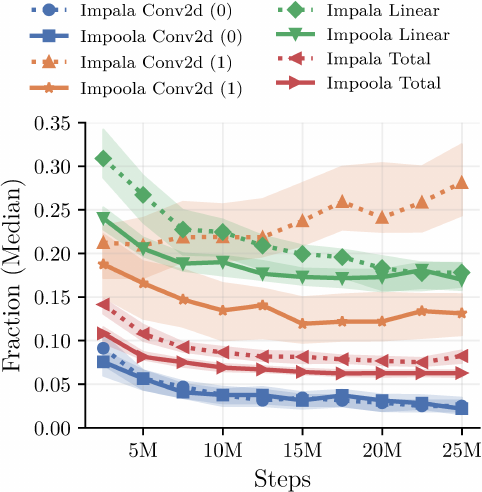}
    \caption{Fraction of dormant neurons per layer throughout training for the first two Conv2d layers of the encoder ($\tau=2$), its output Linear layer, and for the total network. 
    }
    \label{fig:dormant_neurons}
    \end{minipage} 
\end{figure*}

\textbf{Agent-Centered Observations:}
We discuss the influence of translation sensitivity in relation to the characteristics of the Procgen Benchmark games.
The game-specific reward curves are presented in Figure~\ref{fig:ablation}.
First, it can be seen that Impala-CNN only exhibits favorable performance in the Coinrun and Ninja environments, which have agent-centered observations, especially early in training.
However, we find no advantage for Impala-CNN in Caveflyer, despite being an environment with agent-centered observations, and the training curve in Coinrun indicates overfitting.
As shown in Figure~\ref{fig:agent_centered_envs_generalization}, the advantage of Impala-CNN in environments with agent-centered observations diminishes in the \textit{hard} setting.
Impoola-CNN achieves comparable final performance for these environments as the generalization performance of Impala-CNN saturates early.
We reason that positional information may only be a helpful inductive bias early in training for agent-centered games.
However, it is also prone to overfitting, leading to networks that show weaker generalization than networks that pose low translation sensitivity.
The results for AvgPool(2,2) highlight an interesting trade-off as seen in Figure~\ref{fig:ablation}.
While falling short of Impoola-CNN in non-agent-centered games, it outperforms Impala-CNN for them and meets Impala-CNN's performance in agent-centered games.

\textbf{Dormant Neurons:}
When monitoring the fraction of dormant neurons \citep{sokar2023dormant} during \gls*{ppo} training in the \textit{generalization} track (\textit{easy}), we see in Figure~\ref{fig:dormant_neurons} that the total number of dormant neurons decreases during training for both encoder models.
While the Linear layer in Impala-CNN has a higher initial dormant fraction, it decreases during training to the same fraction.
However, we find a distinct difference for the \textit{second} Conv2d layer (1), which has no residual connection such as the first Conv2d layer (0): while Impoola-CNN's count decreases, Impala-CNN's dormant neuron count increases here during training. 
We reason that this may be attributed to better training stability and gradients to this early network layer in Impoola-CNN, as the Impala-CNN unbalanced distribution of network weights along the network depths might reduce effective gradient flow.
\section{Conclusion and Future Work}\label{sec:conclusion}
This work introduces the Impoola-CNN model, an improved image encoder for \gls*{drl} that is based on the widely used Impala-CNN architecture.
Our advancement is based on the introduction of a \gls*{gap} layer to the Impala-CNN, which has a two-fold implication.
First, it leads to a reduction of required weights in the encoder's Linear layer and creates a balanced weight distribution along the network's depth.
Second, we find that this change effectively reduces the network's translation sensitivity.
Our experiments for the full Procgen Benchmark show that Impoola-CNN leads to a significant performance increase, most prominent in environments without agent-centered observations.
We also find that the stronger dependence on absolute positions of Impala-CNN may become detrimental during the longer training for the hard setting.
We hypothesize that its \gls*{gap} layer encourages the agent to learn more universal feature representations, which enable better adaptation to new levels and facilitate learning under challenging conditions.

For future work, we plan to conduct further experiments outside the Procgen Benchmark.
While results for other game-inspired environments, e.g., Atari games, would double-down our results, we see a stronger need for evaluation in real-world image-based \gls*{drl} applications, e.g., automated driving \citep{trumpp2023efficient} or vision-guided quadrupedal locomotion \citep{yang2021learning}.






\subsubsection*{Acknowledgments}
\label{sec:ack}
Marco Caccamo was supported by an Alexander von Humboldt Professorship endowed by the German Federal Ministry of Education and Research.


\bibliography{references}

\begin{thebibliography}{46}
\providecommand{\natexlab}[1]{#1}
\providecommand{\url}[1]{\texttt{#1}}
\expandafter\ifx\csname urlstyle\endcsname\relax
  \providecommand{\doi}[1]{DOI: #1}\else
  \providecommand{\doi}{DOI: \begingroup \urlstyle{rm}\Url}\fi

\bibitem[Agarwal et~al.(2021)Agarwal, Schwarzer, Castro, Courville, and Bellemare]{agarwal2021deep}
Rishabh Agarwal, Max Schwarzer, Pablo~Samuel Castro, Aaron~C Courville, and Marc Bellemare.
\newblock Deep reinforcement learning at the edge of the statistical precipice.
\newblock \emph{Advances in neural information processing systems}, 34:\penalty0 29304--29320, 2021.

\bibitem[Bjorck et~al.(2021)Bjorck, Gomes, and Weinberger]{bjorck2021towards}
Nils Bjorck, Carla~P Gomes, and Kilian~Q Weinberger.
\newblock Towards deeper deep reinforcement learning with spectral normalization.
\newblock \emph{Advances in neural information processing systems}, 34:\penalty0 8242--8255, 2021.

\bibitem[Cobbe et~al.(2019)Cobbe, Klimov, Hesse, Kim, and Schulman]{cobbe2019quantifying}
Karl Cobbe, Oleg Klimov, Chris Hesse, Taehoon Kim, and John Schulman.
\newblock Quantifying generalization in reinforcement learning.
\newblock In \emph{International conference on machine learning}, pp.\  1282--1289. PMLR, 2019.

\bibitem[Cobbe et~al.(2020)Cobbe, Hesse, Hilton, and Schulman]{cobbe2020leveraging}
Karl Cobbe, Chris Hesse, Jacob Hilton, and John Schulman.
\newblock Leveraging procedural generation to benchmark reinforcement learning.
\newblock In \emph{International conference on machine learning}, pp.\  2048--2056. PMLR, 2020.

\bibitem[D'Oro et~al.(2022)D'Oro, Schwarzer, Nikishin, Bacon, Bellemare, and Courville]{d2022sample}
Pierluca D'Oro, Max Schwarzer, Evgenii Nikishin, Pierre-Luc Bacon, Marc~G Bellemare, and Aaron Courville.
\newblock Sample-efficient reinforcement learning by breaking the replay ratio barrier.
\newblock In \emph{Deep Reinforcement Learning Workshop NeurIPS 2022}, 2022.

\bibitem[Espeholt et~al.(2018)Espeholt, Soyer, Munos, Simonyan, Mnih, Ward, Doron, Firoiu, Harley, Dunning, et~al.]{espeholt2018impala}
Lasse Espeholt, Hubert Soyer, Remi Munos, Karen Simonyan, Vlad Mnih, Tom Ward, Yotam Doron, Vlad Firoiu, Tim Harley, Iain Dunning, et~al.
\newblock Impala: Scalable distributed deep-rl with importance weighted actor-learner architectures.
\newblock In \emph{International conference on machine learning}, pp.\  1407--1416. PMLR, 2018.

\bibitem[Funk et~al.(2022)Funk, Chalvatzaki, Belousov, and Peters]{funk2022learn2assemble}
Niklas Funk, Georgia Chalvatzaki, Boris Belousov, and Jan Peters.
\newblock Learn2assemble with structured representations and search for robotic architectural construction.
\newblock In \emph{Conference on Robot Learning}, pp.\  1401--1411. PMLR, 2022.

\bibitem[He et~al.(2016)He, Zhang, Ren, and Sun]{he2016deep}
Kaiming He, Xiangyu Zhang, Shaoqing Ren, and Jian Sun.
\newblock Deep residual learning for image recognition.
\newblock In \emph{Proceedings of the IEEE conference on computer vision and pattern recognition}, pp.\  770--778, 2016.

\bibitem[Henderson et~al.(2018)Henderson, Islam, Bachman, Pineau, Precup, and Meger]{henderson2018deep}
Peter Henderson, Riashat Islam, Philip Bachman, Joelle Pineau, Doina Precup, and David Meger.
\newblock Deep reinforcement learning that matters.
\newblock In \emph{Proceedings of the AAAI conference on artificial intelligence}, volume~32, 2018.

\bibitem[Hu et~al.(2018)Hu, Shen, and Sun]{hu2018squeeze}
Jie Hu, Li~Shen, and Gang Sun.
\newblock Squeeze-and-excitation networks.
\newblock In \emph{Proceedings of the IEEE conference on computer vision and pattern recognition}, pp.\  7132--7141, 2018.

\bibitem[Huang et~al.(2017)Huang, Liu, Van Der~Maaten, and Weinberger]{huang2017densely}
Gao Huang, Zhuang Liu, Laurens Van Der~Maaten, and Kilian~Q Weinberger.
\newblock Densely connected convolutional networks.
\newblock In \emph{Proceedings of the IEEE conference on computer vision and pattern recognition}, pp.\  4700--4708, 2017.

\bibitem[Huang et~al.(2022)Huang, Dossa, Ye, Braga, Chakraborty, Mehta, and Araújo]{huang2022cleanrl}
Shengyi Huang, Rousslan Fernand~Julien Dossa, Chang Ye, Jeff Braga, Dipam Chakraborty, Kinal Mehta, and João~G.M. Araújo.
\newblock Cleanrl: High-quality single-file implementations of deep reinforcement learning algorithms.
\newblock \emph{Journal of Machine Learning Research}, 23\penalty0 (274):\penalty0 1--18, 2022.

\bibitem[Islam* et~al.(2020)Islam*, Jia*, and Bruce]{islam2020much}
Md~Amirul Islam*, Sen Jia*, and Neil D.~B. Bruce.
\newblock How much position information do convolutional neural networks encode?
\newblock In \emph{International Conference on Learning Representations}, 2020.

\bibitem[Kaplan et~al.(2020)Kaplan, McCandlish, Henighan, Brown, Chess, Child, Gray, Radford, Wu, and Amodei]{kaplan2020scaling}
Jared Kaplan, Sam McCandlish, Tom Henighan, Tom~B Brown, Benjamin Chess, Rewon Child, Scott Gray, Alec Radford, Jeffrey Wu, and Dario Amodei.
\newblock Scaling laws for neural language models.
\newblock \emph{arXiv preprint arXiv:2001.08361v1}, 2020.

\bibitem[Kauderer-Abrams(2017)]{kauderer2017quantifying}
Eric Kauderer-Abrams.
\newblock Quantifying translation-invariance in convolutional neural networks.
\newblock \emph{arXiv preprint arXiv:1801.01450v1}, 2017.

\bibitem[Kayhan \& Gemert(2020)Kayhan and Gemert]{kayhan2020translation}
Osman~Semih Kayhan and Jan C~van Gemert.
\newblock On translation invariance in cnns: Convolutional layers can exploit absolute spatial location.
\newblock In \emph{Proceedings of the IEEE/CVF Conference on Computer Vision and Pattern Recognition}, pp.\  14274--14285, 2020.

\bibitem[Kostrikov et~al.(2020)Kostrikov, Yarats, and Fergus]{kostrikov2020image}
Ilya Kostrikov, Denis Yarats, and Rob Fergus.
\newblock Image augmentation is all you need: Regularizing deep reinforcement learning from pixels.
\newblock \emph{arXiv preprint arXiv:2004.13649v4}, 2020.

\bibitem[Lee et~al.(2024)Lee, Hwang, Kim, Kim, Tai, Subramanian, Wurman, Choo, Stone, and Seno]{lee2024simba}
Hojoon Lee, Dongyoon Hwang, Donghu Kim, Hyunseung Kim, Jun~Jet Tai, Kaushik Subramanian, Peter~R Wurman, Jaegul Choo, Peter Stone, and Takuma Seno.
\newblock Simba: Simplicity bias for scaling up parameters in deep reinforcement learning.
\newblock \emph{arXiv preprint arXiv:2410.09754v1}, 2024.

\bibitem[Lee et~al.(2020)Lee, Lee, Shin, and Lee]{lee2020network}
Kimin Lee, Kibok Lee, Jinwoo Shin, and Honglak Lee.
\newblock Network randomization: A simple technique for generalization in deep reinforcement learning.
\newblock In \emph{International Conference on Learning Representations}, 2020.

\bibitem[Lin(2013)]{lin2014network}
M~Lin.
\newblock Network in network.
\newblock \emph{arXiv preprint arXiv:1312.4400v3}, 2013.

\bibitem[Liu et~al.(2022)Liu, Mao, Wu, Feichtenhofer, Darrell, and Xie]{liu2022convnet}
Zhuang Liu, Hanzi Mao, Chao-Yuan Wu, Christoph Feichtenhofer, Trevor Darrell, and Saining Xie.
\newblock A convnet for the 2020s.
\newblock In \emph{Proceedings of the IEEE/CVF conference on computer vision and pattern recognition}, pp.\  11976--11986, 2022.

\bibitem[Lyle et~al.(2019)Lyle, Kwiatkowksa, and Gal]{lyle2019analysis}
Clare Lyle, Marta Kwiatkowksa, and Yarin Gal.
\newblock An analysis of the effect of invariance on generalization in neural networks.
\newblock In \emph{International conference on machine learning Workshop on Understanding and Improving Generalization in Deep Learning}, volume~1, 2019.

\bibitem[Mnih et~al.(2015)Mnih, Kavukcuoglu, Silver, Rusu, Veness, Bellemare, Graves, Riedmiller, Fidjeland, Ostrovski, et~al.]{mnih2015human}
Volodymyr Mnih, Koray Kavukcuoglu, David Silver, Andrei~A Rusu, Joel Veness, Marc~G Bellemare, Alex Graves, Martin Riedmiller, Andreas~K Fidjeland, Georg Ostrovski, et~al.
\newblock Human-level control through deep reinforcement learning.
\newblock \emph{nature}, 518\penalty0 (7540):\penalty0 529--533, 2015.

\bibitem[Mouton et~al.(2020)Mouton, Myburgh, and Davel]{mouton2020stride}
Coenraad Mouton, Johannes~C Myburgh, and Marelie~H Davel.
\newblock Stride and translation invariance in cnns.
\newblock In \emph{Southern African Conference for Artificial Intelligence Research}, pp.\  267--281. Springer, 2020.

\bibitem[Nauman et~al.(2024)Nauman, Ostaszewski, Jankowski, Mi{\l}o{\'s}, and Cygan]{nauman2024bigger}
Michal Nauman, Mateusz Ostaszewski, Krzysztof Jankowski, Piotr Mi{\l}o{\'s}, and Marek Cygan.
\newblock Bigger, regularized, optimistic: scaling for compute and sample-efficient continuous control.
\newblock In \emph{ICML 2024 Workshop: Aligning Reinforcement Learning Experimentalists and Theorists}, 2024.

\bibitem[Nikishin et~al.(2022)Nikishin, Schwarzer, D’Oro, Bacon, and Courville]{nikishin2022primacy}
Evgenii Nikishin, Max Schwarzer, Pierluca D’Oro, Pierre-Luc Bacon, and Aaron Courville.
\newblock The primacy bias in deep reinforcement learning.
\newblock In \emph{International conference on machine learning}, pp.\  16828--16847. PMLR, 2022.

\bibitem[Obando-Ceron et~al.(2024{\natexlab{a}})Obando-Ceron, Courville, and Castro]{obando2024deep}
Johan~Samir Obando-Ceron, Aaron Courville, and Pablo~Samuel Castro.
\newblock In value-based deep reinforcement learning, a pruned network is a good network.
\newblock In \emph{Forty-first International Conference on Machine Learning}, 2024{\natexlab{a}}.

\bibitem[Obando-Ceron et~al.(2024{\natexlab{b}})Obando-Ceron, Sokar, Willi, Lyle, Farebrother, Foerster, Dziugaite, Precup, and Castro]{ceron2024mixtures}
Johan~Samir Obando-Ceron, Ghada Sokar, Timon Willi, Clare Lyle, Jesse Farebrother, Jakob~Nicolaus Foerster, Gintare~Karolina Dziugaite, Doina Precup, and Pablo~Samuel Castro.
\newblock Mixtures of experts unlock parameter scaling for deep {RL}.
\newblock In \emph{Forty-first International Conference on Machine Learning}, 2024{\natexlab{b}}.

\bibitem[Paszke et~al.(2017)Paszke, Gross, Chintala, Chanan, Yang, DeVito, Lin, Desmaison, Antiga, and Lerer]{paszke2017automatic}
Adam Paszke, Sam Gross, Soumith Chintala, Gregory Chanan, Edward Yang, Zachary DeVito, Zeming Lin, Alban Desmaison, Luca Antiga, and Adam Lerer.
\newblock Automatic differentiation in pytorch.
\newblock 2017.

\bibitem[Raileanu \& Fergus(2021)Raileanu and Fergus]{raileanu2021decoupling}
Roberta Raileanu and Rob Fergus.
\newblock Decoupling value and policy for generalization in reinforcement learning.
\newblock In \emph{International Conference on Machine Learning}, pp.\  8787--8798. PMLR, 2021.

\bibitem[Raileanu et~al.(2021)Raileanu, Goldstein, Yarats, Kostrikov, and Fergus]{raileanu2021automatic}
Roberta Raileanu, Maxwell Goldstein, Denis Yarats, Ilya Kostrikov, and Rob Fergus.
\newblock Automatic data augmentation for generalization in reinforcement learning.
\newblock \emph{Advances in Neural Information Processing Systems}, 34:\penalty0 5402--5415, 2021.

\bibitem[Schaul et~al.(2015)Schaul, Quan, Antonoglou, and Silver]{schaul2015prioritized}
Tom Schaul, John Quan, Ioannis Antonoglou, and David Silver.
\newblock Prioritized experience replay.
\newblock \emph{CoRR}, abs/1511.05952, 2015.

\bibitem[Schulman et~al.(2015)Schulman, Moritz, Levine, Jordan, and Abbeel]{schulman2015high}
John Schulman, Philipp Moritz, Sergey Levine, Michael~I. Jordan, and P.~Abbeel.
\newblock High-dimensional continuous control using generalized advantage estimation.
\newblock \emph{CoRR}, abs/1506.02438, 2015.

\bibitem[Schulman et~al.(2017)Schulman, Wolski, Dhariwal, Radford, and Klimov]{schulman2017proximal}
John Schulman, Filip Wolski, Prafulla Dhariwal, Alec Radford, and Oleg Klimov.
\newblock Proximal policy optimization algorithms.
\newblock \emph{arXiv preprint arXiv:1707.06347v2}, 2017.

\bibitem[Schwarzer et~al.(2023)Schwarzer, Ceron, Courville, Bellemare, Agarwal, and Castro]{schwarzer2023bigger}
Max Schwarzer, Johan Samir~Obando Ceron, Aaron Courville, Marc~G Bellemare, Rishabh Agarwal, and Pablo~Samuel Castro.
\newblock Bigger, better, faster: Human-level atari with human-level efficiency.
\newblock In \emph{International Conference on Machine Learning}, pp.\  30365--30380. PMLR, 2023.

\bibitem[Sinha et~al.(2020)Sinha, Bharadhwaj, Srinivas, and Garg]{sinha2020d2rl}
Samarth Sinha, Homanga Bharadhwaj, Aravind Srinivas, and Animesh Garg.
\newblock D2rl: Deep dense architectures in reinforcement learning.
\newblock \emph{arXiv preprint arXiv:2010.09163v2}, 2020.

\bibitem[Sokar et~al.(2023)Sokar, Agarwal, Castro, and Evci]{sokar2023dormant}
Ghada Sokar, Rishabh Agarwal, Pablo~Samuel Castro, and Utku Evci.
\newblock The dormant neuron phenomenon in deep reinforcement learning.
\newblock In \emph{International Conference on Machine Learning}, pp.\  32145--32168. PMLR, 2023.

\bibitem[Sokar et~al.(2024)Sokar, Obando-Ceron, Courville, Larochelle, and Castro]{sokar2024don}
Ghada Sokar, Johan Obando-Ceron, Aaron Courville, Hugo Larochelle, and Pablo~Samuel Castro.
\newblock Don't flatten, tokenize! unlocking the key to softmoe's efficacy in deep {RL}.
\newblock \emph{arXiv preprint arXiv:2410.01930v1}, 2024.

\bibitem[Song et~al.(2020)Song, Jiang, Tu, Du, and Neyshabur]{song2020observational}
Xingyou Song, Yiding Jiang, Stephen Tu, Yilun Du, and Behnam Neyshabur.
\newblock Observational overfitting in reinforcement learning.
\newblock In \emph{International Conference on Learning Representations}, 2020.

\bibitem[Sutton(1988)]{sutton1988learning}
Richard~S Sutton.
\newblock Learning to predict by the methods of temporal differences.
\newblock \emph{Machine learning}, 3:\penalty0 9--44, 1988.

\bibitem[Trumpp et~al.(2023)Trumpp, Büchner, Valada, and Caccamo]{trumpp2023efficient}
Raphael Trumpp, Martin Büchner, Abhinav Valada, and Marco Caccamo.
\newblock Efficient learning of urban driving policies using bird's eye-view state representations.
\newblock In \emph{2023 IEEE 26th International Conference on Intelligent Transportation Systems (ITSC)}, pp.\  4181--4186, 2023.

\bibitem[Van~Hasselt et~al.(2016)Van~Hasselt, Guez, and Silver]{van2016deep}
Hado Van~Hasselt, Arthur Guez, and David Silver.
\newblock Deep reinforcement learning with double q-learning.
\newblock In \emph{Proceedings of the AAAI conference on artificial intelligence}, volume~30, 2016.

\bibitem[Xie et~al.(2017)Xie, Girshick, Doll{\'a}r, Tu, and He]{xie2017aggregated}
Saining Xie, Ross Girshick, Piotr Doll{\'a}r, Zhuowen Tu, and Kaiming He.
\newblock Aggregated residual transformations for deep neural networks.
\newblock In \emph{Proceedings of the IEEE conference on computer vision and pattern recognition}, pp.\  1492--1500, 2017.

\bibitem[Yang et~al.(2021)Yang, Zhang, Hansen, Xu, and Wang]{yang2021learning}
Ruihan Yang, Minghao Zhang, Nicklas Hansen, Huazhe Xu, and Xiaolong Wang.
\newblock Learning vision-guided quadrupedal locomotion end-to-end with cross-modal transformers.
\newblock In \emph{Deep RL Workshop NeurIPS}, 2021.

\bibitem[Zhai et~al.(2022)Zhai, Kolesnikov, Houlsby, and Beyer]{zhai2022scaling}
Xiaohua Zhai, Alexander Kolesnikov, Neil Houlsby, and Lucas Beyer.
\newblock Scaling vision transformers.
\newblock In \emph{Proceedings of the IEEE/CVF conference on computer vision and pattern recognition}, pp.\  12104--12113, 2022.

\bibitem[Zhang et~al.(2018)Zhang, Vinyals, Munos, and Bengio]{zhang2018study}
Chiyuan Zhang, Oriol Vinyals, Remi Munos, and Samy Bengio.
\newblock A study on overfitting in deep reinforcement learning.
\newblock \emph{arXiv preprint arXiv:1804.06893v2}, 2018.

\end{thebibliography}
\bibliographystyle{rlj}

\beginSupplementaryMaterials


\renewcommand{\thesection}{\Alph{section}} 
\renewcommand{\thetable}{\thesection.\arabic{table}} 
\renewcommand{\thefigure}{\thesection.\arabic{figure}} 
\renewcommand{\theequation}{\thesection.\arabic{equation}} 

\setcounter{section}{0}
\setcounter{table}{0} 
\setcounter{figure}{0} 
\setcounter{equation}{0} 



\appendix

\section{Procgen Benchmark}
\label{ap:sec:procgen}
\begin{figure}[!h]
    \centering
    \hspace{0.15\textwidth}
    \hfill
    \begin{subfigure}{0.15\textwidth}
        \includegraphics[width=\textwidth]{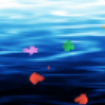}
    \end{subfigure}
    \hfill
    \begin{subfigure}{0.15\textwidth}
        \includegraphics[width=\textwidth]{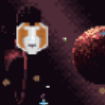}
    \end{subfigure}
    \hfill
    \begin{subfigure}{0.15\textwidth}
        \includegraphics[width=\textwidth]{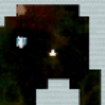}
    \end{subfigure}
    \hfill
    \begin{subfigure}{0.15\textwidth}
        \includegraphics[width=\textwidth]{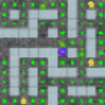}
    \end{subfigure}
    \hspace{0.15\textwidth}
    \hfill
    \\
    \vspace{1em}
    \hspace{0.15\textwidth}
    \hfill
    \begin{subfigure}{0.15\textwidth}
        \includegraphics[width=\textwidth]{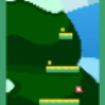}
    \end{subfigure}
    \hfill
    \begin{subfigure}{0.15\textwidth}
        \includegraphics[width=\textwidth]{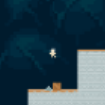}
    \end{subfigure}
    \hfill
    \begin{subfigure}{0.15\textwidth}
        \includegraphics[width=\textwidth]{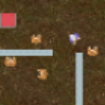}
    \end{subfigure}
    \hfill
    \begin{subfigure}{0.15\textwidth}
        \includegraphics[width=\textwidth]{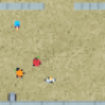}
    \end{subfigure}
    \hspace{0.15\textwidth}
    \hfill
    \\
    \vspace{1em}
    \hspace{0.15\textwidth}
    \hfill
    \begin{subfigure}{0.15\textwidth}
        \includegraphics[width=\textwidth]{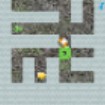}
    \end{subfigure}
    \hfill
    \begin{subfigure}{0.15\textwidth}
        \includegraphics[width=\textwidth]{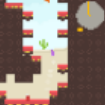}
    \end{subfigure}
    \hfill
    \begin{subfigure}{0.15\textwidth}
        \includegraphics[width=\textwidth]{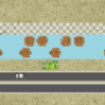}
    \end{subfigure}
    \hfill
    \begin{subfigure}{0.15\textwidth}
        \includegraphics[width=\textwidth]{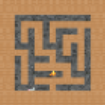}
    \end{subfigure}
    \hspace{0.15\textwidth}
    \hfill
    \\
    \vspace{1em} 
    \hspace{0.15\textwidth}
    \hfill
    \begin{subfigure}{0.15\textwidth}
        \includegraphics[width=\textwidth]{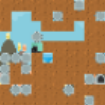}
    \end{subfigure}
    \hfill
    \begin{subfigure}{0.15\textwidth}
        \includegraphics[width=\textwidth]{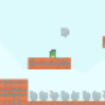}
    \end{subfigure}
    \hfill
    \begin{subfigure}{0.15\textwidth}
        \includegraphics[width=\textwidth]{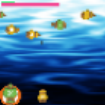}
    \end{subfigure}
    \hfill
    \begin{subfigure}{0.15\textwidth}
        \includegraphics[width=\textwidth]{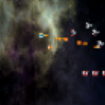}
    \end{subfigure}
    \hspace{0.15\textwidth}
    \hfill
    \\
    \caption{All ProcGen environments depicted with a single image observation (64x64 pixels): Bigfish, Bossfight, Caveflyer, Chaser, Climber, Coinrun, Dodgeball, Fruitbot, Heist, Jumper, Leaper, Maze, Miner, Ninja, Plunder, and Starpilot (left to right).}
    \label{fig:all_procgen_games}
\end{figure}

\textbf{Description:} The Procgen Benchmark was developed by \citet{cobbe2020leveraging} to test generalization and sample efficiency of \gls*{drl} agents.
The Benchmark consists of 16 games and allows for \textit{hard} and \textit{easy} game settings to balance computation demand accordingly.
For the \textit{generalization} track, a restricted fixed set of 200 levels is used for training in the \textit{easy} setting, while all possible procedurally generated levels are used for evaluation.
\citet{cobbe2020leveraging} recommend training for 25M steps in this setting.
When the \textit{hard} setting is used, 1000 training levels are used with 100M training steps.
The \textit{efficiency} tracks do not restrict the set of training levels but use the full distribution of levels.
The action space of the Procgen environments consists of 15 discrete actions.
Observations are RGB images with $3\times64\times64$ pixels.
No stacking of images is required, as we utilize the environments without the setting that requires memory.
We show example image observations for the games in Figure~\ref{fig:all_procgen_games} and list specific game characteristics in Table~\ref{tab:characteristics_procgen}.

\textbf{Normalized Score:} As suggested by \cite{cobbe2020leveraging}, we report normalized scores $S$ by

\begin{equation}
\label{eq:normalized_scores}
    S = \frac{R - R_{\text{min}}}{R_{\text{max}} - R_{\text{min}}},
\end{equation}
where \( R \) is the raw return collected by the agent, \( R_{\text{min}} \) is the score for the environment by a random agent, \( R_{\text{max}} \) is the maximum possible score.
The normalization constants are shown in Table~\ref{tab:normalization_constants_easy} and Table~\ref{tab:normalization_constants_hard} for \textit{easy} and \textit{hard} game settings, respectively.

\begin{table}
\centering
\caption{Game characteristics of the Procgen Benchmark environments. Fixed translation in the x and y directions means the image is centered on the agent, i.e., there is no relative movement of the agent in the image. Agent-centered images convey that the map observation is not fixed but moves relatively to the agent.}
\label{tab:characteristics_procgen}
\begin{tabular}{l l l l l}
\toprule
\textbf{Game} &  \textbf{X Translation}  & \textbf{Y Translation} &  \textbf{Rotation} & \textbf{Map}\\
\midrule
Bigfish & Free & Free & Left/right & Fixed \\
Bossfight & Free & Limited & No & Fixed \\
Caveflyer & Fixed & Fixed & Free & Free \\
Chaser & Free & Free & No & Fixed \\
Climber & Free & Fixed & Left/right & Fixed  \\
Coinrun & Fixed & Fixed & No & Free \\
Dodgeball & Free & Free & Free & Fixed \\
Fruitbot & Free & Fixed & No & Free \\
Heist & Free & Free & No & Fixed \\
Jumper & Fixed & Fixed & Left/right & Free \\
Leaper & Free & Free & Free & Fixed \\
Maze & Free & Free & Free & Fixed \\
Miner & Free & Free & Fixed & Fixed \\
Ninja & Fixed & Fixed & Left/right & Free \\
Plunder & Free & Fixed & No & Fixed \\
Starpilot & Free & Free & Free & Free \\
\bottomrule
\end{tabular}
\end{table}

\begin{table}
\centering
\caption{Normalization constants from \cite{cobbe2020leveraging} for all Procgen Benchmark environments in the \textit{easy} setting.}
\label{tab:normalization_constants_easy}
\begin{tabular}{c c c | c c c}
\toprule
\textbf{Game} &  $\boldsymbol{R}_{\textbf{min}}$ &  $\boldsymbol{R}_{\textbf{max}}$ & \textbf{Game} &  $\boldsymbol{R}_{\textbf{min}}$ &  $\boldsymbol{R}_{\textbf{max}}$ \\
\midrule
Bigfish & 1 & 40 & Jumper & 3 & 10 \\
Bossfight & 0.5  & 13 & Leaper & 3 & 10 \\
Caveflyer & 3.5  & 12 & Maze & 5 & 10 \\
Chaser & 0.5 & 13 & Miner & 1.5 & 13 \\
Climber & 2 & 12.6 & Ninja & 3.5 & 10 \\
Coinrun & 5 & 10 & Plunder & 4.5 & 30 \\
Dodgeball & 1.5 & 19 & Starpilot & 2.5 & 64 \\
Fruitbot & -1.5 & 32.4 & Heist & 3.5 & 10 \\
\bottomrule
\end{tabular}
\end{table}

\begin{table}
\centering
\caption{Normalization constants from \cite{cobbe2020leveraging} for all Procgen Benchmark environments in the \textit{hard} setting.}
\label{tab:normalization_constants_hard}
\begin{tabular}{c c c | c c c}
\toprule
\textbf{Game} &  $\boldsymbol{R}_{\textbf{min}}$ &  $\boldsymbol{R}_{\textbf{max}}$ & \textbf{Game} &  $\boldsymbol{R}_{\textbf{min}}$ &  $\boldsymbol{R}_{\textbf{max}}$ \\
\midrule
Bigfish & 0 & 40 & Jumper & 1 & 10 \\
Bossfight & 0.5  & 13 & Leaper & 1.5 & 10 \\
Caveflyer & 2  & 13.4 & Maze & 4 & 10 \\
Chaser & 0.5 & 14.2 & Miner & 1.5 & 20 \\
Climber & 1 & 12.6 & Ninja & 2 & 10 \\
Coinrun & 5 & 10 & Plunder & 3 & 30 \\
Dodgeball & 1.5 & 19 & Starpilot & 1.5 & 35 \\
Fruitbot & -0.5 & 27.2 & Heist & 2 & 10 \\
\bottomrule
\end{tabular}
\end{table}

\clearpage
\section{Experiment Details}

\subsection{Hyperparameters List}
\label{ap:sec:hyperparameters}

\begin{table*}[!h]
\centering
\caption{Hyperparameters for Proximal Policy Optimization (PPO).}
\label{tab:ppo_hyperparameters}

\begin{tabular}{p{6cm} p{3cm}}
\toprule
\textbf{Hyperparameter} & \textbf{Values} \\
\midrule
Number Parallel Environments & 64 \\
Environment Steps  & 256 \\
Learning Rate ($\tau=2$) &  $3.5 \times 10^{-4}$ \\
Batch Size & 2048 \\
Epochs &  3 \\
Discount Factor $\gamma$ & 0.99 \\
GAE Lambda ($\lambda$) &  0.95 \\
Clip Range & 0.2 \\
{Value Function Coefficient} & 0.5 \\
Entropy Coefficient &  0.01 \\
Max Gradient Norm  & 0.5 \\
Optimizer & Adam \\
Shared Policy and Value Network  & Yes \\
\bottomrule
\end{tabular}

\end{table*}

\begin{table*}[!h]
\centering

\caption{Hyperparameters for Deep Q-Network (DQN).}
\label{tab:dqn_hyperparameters}

\begin{tabular}{p{6cm} p{3cm}}
\toprule
\textbf{Hyperparameter} & \textbf{Values} \\
\midrule
Number Parallel Environments & 128 \\
Learning Rate ($\tau=2$) & $1 \times 10^{-4}$ \\
Batch Size & 512 \\
Discount Factor $\gamma$ & 0.99 \\
Target Network Update Frequency & 64,000 steps \\
Learning Starts & 250,000 steps \\
Train Frequency & 1 \\
Replay Buffer Size & $1 \times 10^6$ \\
Exploration Initial $\epsilon$ & 1.0 \\
Exploration Final $\epsilon$ & 0.025 \\
Exploration Decay Fractions & 0.1 \\
Max Gradient Norm  & 10.0 \\
Optimizer & Adam \\
\bottomrule
\end{tabular}

\end{table*}

\subsection{Benchmark Methods}\label{ap:sec:benchmark_methods}
\begin{itemize}
    \item Unstructured gradual magnitude pruning \citep{obando2024deep}: PyTorch provides tools for unstructured pruning. We use a target sparsity of 0.9 and follow the proposed schedule that starts pruning 20~\% into training and stops at 80~\%.
    \item \Gls*{softmoe} \citep{ceron2024mixtures}: PyTorch reimplementation on base of the official code release \url{https://github.com/google/dopamine/tree/master/dopamine/labs/moes}. We deploy 10 experts.
    \item ReDo \citep{sokar2023dormant}: PyTorch reimplementation from \url{https://github.com/timoklein/redo} that was based on the official code release. We initialize neurons every 100 iterations and set $\tau=0.025$.
\end{itemize}

\newpage

\subsection{Translation Sensitivity Maps}\label{sec:translation_sensitivity_maps_appendix}

\begin{figure}[h]
    \centering
    \includegraphics[width=0.8\textwidth]{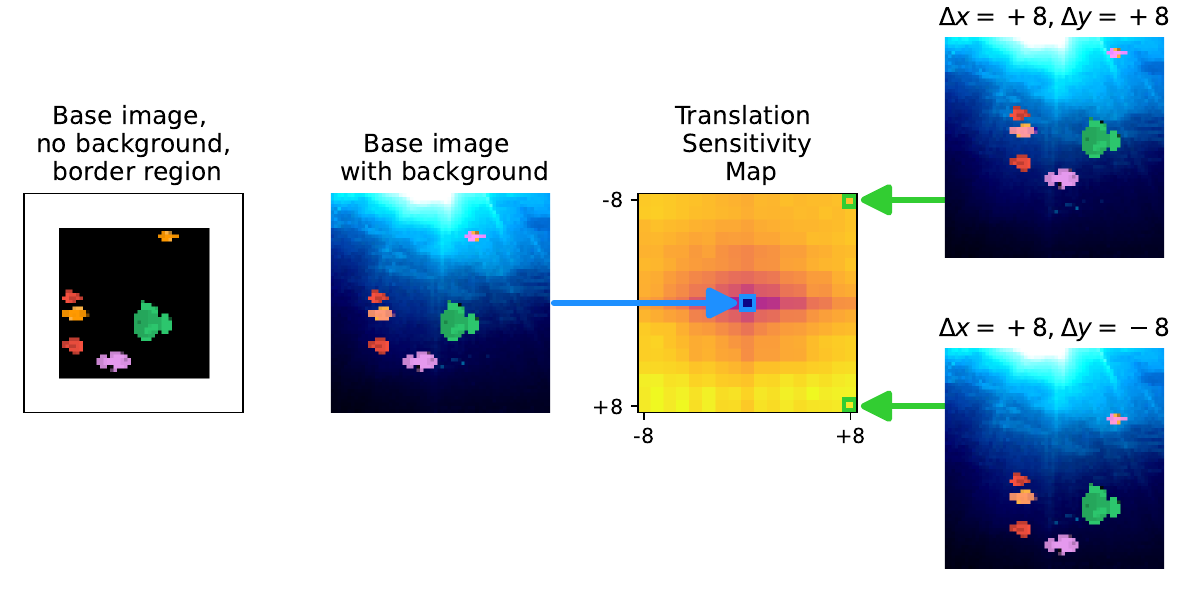}
    \caption{
    Generation of translation sensitivity maps.
    For the arrangement of fish, we require a free border region (white) with a width of $10$ pixels, which ensures that translations by $\pm 8$ pixels do not remove any fish from the frame (\textbf{left}).
    We then add the background to obtain the base image, which provides the base actor output (\textbf{middle left)}.
    Each pixel of the translation sensitivity map
    (\textbf{middle right})
    corresponds to a translation 
    of the entities in the foreground
    relative to the base image. 
    For example, the pixel in the right upper corner corresponds to the maximal translation to the right and upward (\textbf{upper right}).
    }
    \label{fig:translation_sensitivity_explaination}
\end{figure}

\textbf{Adapting Translation Sensitivity Maps for Procgen:}
Originally, translation sensitivity maps were generated by shifting MNIST digits, which feature unicolor backgrounds.
In this case, the backgrounds fuse naturally and no artifacts are introduced.
This approach cannot be transferred to Procgen as the backgrounds are not unicolor. 
Moreover, it is more meaningful to simply measure the agent's sensitivity to translations of the entities in the foreground, while keeping the original $64 \times 64$ pixel background.
Translation sensitivity maps can then be created as a visualization where each pixel (x,y) of a heatmap corresponds to the translation sensitivity score $s$ given an image $x_\text{trans}$ that was translated by (x,y) pixels compared to an original image $x_\text{orig}$.
As shown in Figure~\ref{fig:translation_sensitivity_explaination}, the x and y axes in these maps are centered around 0 which means the center pixel's sensitivity score is always 0 as it references the untranslated image.

\textbf{Data Generation:}
We select the Bigfish game for evaluation and data generation, as its Procgen game engine allows us to create images of the background and fish entities independently.
We originate data from different episodes to ensure data diversity and constrain the agent to be in the center $22\times22$ pixels square of the frame to mitigate the effect of proximity to the image border.
We also ensure that the agent is not alone in the frame, so only interactions with other fish are investigated.
As shown in Figure~\ref{fig:translation_sensitivity_explaination}, the collected fish images can then be translated independently over different background scenarios to create translated images $x_\text{trans}$ given the original image $x_\text{orig}$.

\textbf{Measuring Translation Sensitivity}
The actor network $\textnormal{Actor}_{\theta}$ in \gls*{ppo} with discrete actions outputs the logits of a Categorical distribution, from which actions \( a \in \{1, \dots, 15\} \) are sampled.  
We define the translation sensitivity score $s$ as follows
\begin{equation}
    s = \big\| \text{SoftMax}(\textnormal{Actor}_{\theta}(x_\text{orig})) - \text{SoftMax}(\textnormal{Actor}_{\theta}(x_\text{trans})) \big\|_1,
\end{equation}
quantifying how much the action probabilities change when translating the input image $x_\text{orig}$ to $x_\text{trans}$ while not altering the relative positioning of the entities.
Unlike the approach of \citet{kauderer2017quantifying} that measures translation sensitivity by computing distances of the network's output vector and then normalizes the score, our method directly operates in the action probability space.
By computing the L1 distance between probability distributions, our method ensures a common probability space that directly allows for comparisons between different networks, in contrast to the normalization of the output vector, where finding a meaningful normalization is not obvious.

\newpage
\section{Network Architecture}
\label{ap:subsec:detailed_architectures}

\begin{table}[!h]
\centering
\scriptsize

\caption{Model summary of the Impala-CNN ($\tau=2$) for \gls*{ppo} with 64 x 64 input images. The overall parameter count is 1,441,680, with a total of 118.26M multi-adds.}
\label{ap:tab:impala_net}

\begin{tabularx}{\textwidth}{lXXXXXX}
\toprule
\textbf{Layer (type:depth-idx)} & \textbf{Input} & \textbf{Output} & \textbf{Param \#} & \textbf{Kernel} & \textbf{Param \%} & \textbf{Multi-Adds} \\
\midrule
ImpalaPPOActorCritic           & [3, 64, 64]   & [15]         & --      & --       & --      & -- \\
\quad Impala-CNN         & [3, 64, 64]   & [256]        & --      & --       & --      & -- \\
\quad \quad ConvSequence    & [3, 64, 64]   & [32, 32, 32] & --      & --       & --      & -- \\
\quad \quad \quad Conv2d     & [3, 64, 64]   & [32, 64, 64] & 896   & [3, 3]   & 0.06\%  & 3,670,016 \\
\quad \quad \quad ResidualBlock & [32, 32, 32] & [32, 32, 32] & --      & --       & --      & -- \\
\quad \quad \quad \quad Conv2d & [32, 32, 32] & [32, 32, 32] & 9,248   & [3, 3]   & 0.64\%  & 9,469,952 \\
\quad \quad \quad \quad Conv2d & [32, 32, 32] & [32, 32, 32] & 9,248   & [3, 3]   & 0.64\%  & 9,469,952 \\
\quad \quad \quad ResidualBlock & [32, 32, 32] & [32, 32, 32] & --      & --       & --      & -- \\
\quad \quad \quad \quad Conv2d & [32, 32, 32] & [32, 32, 32] & 9,248    & [3, 3]   & 0.64\%  & 9,469,952 \\
\quad \quad \quad \quad Conv2d & [32, 32, 32] & [32, 32, 32] & 9,248  & [3, 3]   & 0.64\%  & 9,469,952 \\
\quad \quad ConvSequence   & [32, 32, 32]  & [64, 16, 16] & --      & --       & --      & -- \\
\quad \quad \quad Conv2d     & [32, 32, 32]  & [64, 32, 32] & 18,496  & [3, 3]   &   1.28\%  & 18,939,904 \\
\quad \quad \quad ResidualBlock  & [64, 16, 16] & [64, 16, 16] & --      & --       & --      & -- \\
\quad \quad \quad \quad Conv2d  & [64, 16, 16] & [64, 16, 16] & 36,928  & [3, 3]   & 2.56\%  & 9,453,568 \\
\quad \quad \quad \quad Conv2d & [64, 16, 16] & [64, 16, 16] & 36,928  & [3, 3]   & 2.56\%  & 9,453,568 \\
\quad \quad \quad ResidualBlock & [64, 16, 16] & [64, 16, 16] & --      & --       & --      & -- \\
\quad \quad \quad \quad Conv2d  & [64, 16, 16] & [64, 16, 16] & 36,928  & [3, 3]   & 2.56\%  & 9,453,568 \\
\quad \quad \quad \quad Conv2d  & [64, 16, 16] & [64, 16, 16] & 36,928  & [3, 3]   & 2.56\%  & 9,453,568 \\
\quad \quad ConvSequence     & [64, 16, 16]  & [64, 8, 8]   & --      & --       & --      & -- \\
\quad \quad \quad Conv2d   & [64, 16, 16]  & [64, 16, 16] & 36,928  & [3, 3]   & 2.56\%  & 9,453,568 \\
\quad \quad \quad ResidualBlock & [64, 8, 8]   & [64, 8, 8]   & --      & --       & --      & -- \\
\quad \quad \quad \quad Conv2d & [64, 8, 8]   & [64, 8, 8]   & 36,928  & [3, 3]   & 2.56\%  & 2,363,392 \\
\quad \quad \quad \quad Conv2d & [64, 8, 8]  & [64, 8, 8]   & 36,928  & [3, 3]   & 2.56\%  & 2,363,392 \\
\quad \quad \quad ResidualBlock & [64, 8, 8]  & [64, 8, 8]   & --      & --       & --      & -- \\
\quad \quad \quad \quad Conv2d & [64, 8, 8]  & [64, 8, 8]   & 36,928  & [3, 3]   & 2.56\%  & 2,363,392 \\
\quad \quad \quad \quad Conv2d& [64, 8, 8]  & [64, 8, 8]   & 36,928  & [3, 3]   & 2.56\%  & 2,363,392\\
\quad \quad Flatten& [64, 8, 8]   & [4096]   & --      & --       & --      & -- \\
\quad \quad Linear           & [4096]       & [256]        & 1,048,832  & --    & 72.75\% & 1,048,832 \\
Actor                        & [256]        & [15]         & 3,855   & --       & 0.27\%  & 3,855 \\
Critic                       & [256]        & [1]          & 257     & --       & 0.02\%  & 257 \\
\bottomrule
\end{tabularx}

\end{table}

\begin{table}[!h]
\centering
\scriptsize

\caption{Model summary of the Impoola-CNN ($\tau=2$) for \gls*{ppo} with 64 x 64 input images. The overall parameter count is 409,488, with a total of 117.23M multi-adds.}
\label{ap:tab:impoola_net}

\begin{tabularx}{\textwidth}{lXXXXXX}
\toprule
\textbf{Layer (type:depth-idx)} & \textbf{Input} & \textbf{Output} & \textbf{Param \#} & \textbf{Kernel} & \textbf{Param \%} & \textbf{Multi-Adds} \\
\midrule
ImpoolaPPOActorCritic              & [3, 64, 64]   & [15]         & --      & --       & --      & -- \\
\quad Impoola-CNN           & [3, 64, 64]   & [256]        & --      & --       & --      & -- \\
\quad \quad ConvSequence     & [3, 64, 64]   & [32, 32, 32] & --      & --       & --      & -- \\
\quad \quad \quad Conv2d     & [3, 64, 64]   & [32, 64, 64] & 896   & [3, 3]   & 0.22\%  & 3,670,016 \\
\quad \quad \quad ResidualBlock & [32, 32, 32] & [32, 32, 32] & --      & --       & --      & -- \\
\quad \quad \quad \quad Conv2d & [32, 32, 32] & [32, 32, 32] & 9,248   & [3, 3]   & 2.26\%  & 9,469,952 \\
\quad \quad \quad \quad Conv2d & [32, 32, 32] & [32, 32, 32] & 9,248   & [3, 3]   & 2.26\%  & 9,469,952 \\
\quad \quad \quad ResidualBlock & [32, 32, 32] & [32, 32, 32] & --      & --       & --      & -- \\
\quad \quad \quad \quad Conv2d & [32, 32, 32] & [32, 32, 32] & 9,248    & [3, 3]   & 2.26\%  & 9,469,952 \\
\quad \quad \quad \quad Conv2d & [32, 32, 32] & [32, 32, 32] & 9,248  & [3, 3]   & 2.26\%  & 9,469,952 \\
\quad \quad ConvSequence     & [32, 32, 32]  & [64, 16, 16] & --      & --       & --      & -- \\
\quad \quad \quad Conv2d    & [32, 32, 32]  & [64, 32, 32] & 18,496  & [3, 3]   &  4.52\%  & 18,939,904 \\
\quad \quad \quad ResidualBlock & [64, 16, 16] & [64, 16, 16] & --      & --       & --      & -- \\
\quad \quad \quad \quad Conv2d & [64, 16, 16] & [64, 16, 16] & 36,928  & [3, 3]   & 9.02\%  & 9,453,568 \\
\quad \quad \quad \quad Conv2d& [64, 16, 16] & [64, 16, 16] & 36,928  & [3, 3]   & 9.02\%  & 9,453,568 \\
\quad \quad \quad ResidualBlock & [64, 16, 16] & [64, 16, 16] & --      & --       & --      & -- \\
\quad \quad \quad \quad Conv2d & [64, 16, 16] & [64, 16, 16] & 36,928  & [3, 3]   & 9.02\%  & 9,453,568 \\
\quad \quad \quad \quad Conv2d & [64, 16, 16] & [64, 16, 16] & 36,928  & [3, 3]   & 9.02\%  & 9,453,568 \\
\quad \quad ConvSequence     & [64, 16, 16]  & [64, 8, 8]   & --      & --       & --      & -- \\
\quad \quad \quad Conv2d     & [64, 16, 16]  & [64, 16, 16] & 36,928  & [3, 3]   & 9.02\%  & 9,453,568 \\
\quad \quad \quad ResidualBlock & [64, 8, 8]   & [64, 8, 8]   & --      & --       & --      & -- \\
\quad \quad \quad \quad Conv2d & [64, 8, 8]   & [64, 8, 8]   & 36,928  & [3, 3]   & 9.02\%  & 2,363,392 \\
\quad \quad \quad \quad Conv2d & [64, 8, 8]  & [64, 8, 8]   & 36,928  & [3, 3]   & 9.02\%  & 2,363,392 \\
\quad \quad \quad ResidualBlock & [64, 8, 8]  & [64, 8, 8]   & --      & --       & --      & -- \\
\quad \quad \quad \quad Conv2d & [64, 8, 8]  & [64, 8, 8]   & 36,928  & [3, 3]   & 9.02\%  & 2,363,392 \\
\quad \quad \quad \quad Conv2d & [64, 8, 8]  & [64, 8, 8]   & 36,928  & [3, 3]   & 9.02\%  & 2,363,392\\
\quad \quad AdaptiveAvgPool2d & [64, 8, 8]   & [64, 1, 1]   & --      & --       & --      & -- \\
\quad \quad Linear           & [64]       & [256]        & 16,640  & --    & 4.06\% & 16,640 \\
Actor                       & [256]        & [15]         & 3,855   & --       & 0.94\%  & 3,855 \\
Critic                      & [256]        & [1]          & 257     & --       & 0.06\%  & 257 \\
\bottomrule
\end{tabularx}

\end{table}

\clearpage

\section{Nature-CNN with GAP}\label{app:add_experiments}
As suggested during the reviewing process, we also evaluate the effect of adding \gls*{gap} to the Nature-CNN \citep{mnih2015human}.
This experiment aims to improve the understanding of the effect of adding a \gls*{gap} layer.
However, as the overall performance of the Nature-CNN is weak and there are limited scaling gains, the results are not fully conclusive.

\begin{figure}[h]
  \centering
  \begin{tikzpicture}
    \node[anchor=south west, inner sep=0] (grid) at (0,0) 
      {
        \includegraphics[width=1\textwidth]{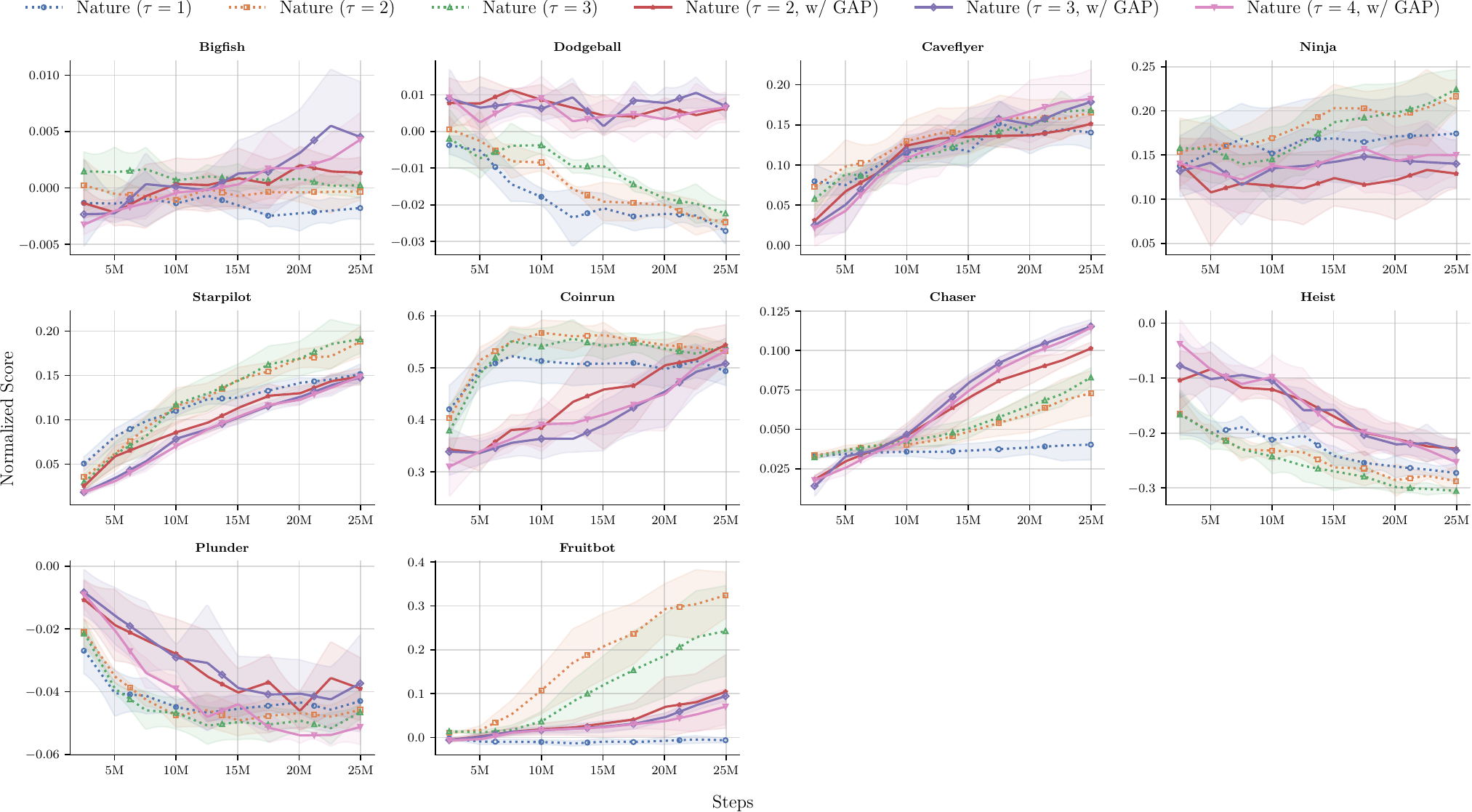}
      };
    
    \begin{scope}[x={(grid.south east)}, y={(grid.north west)}]
      \node[anchor=south west] at (0.52,0) 
        {
        \includegraphics[width=0.45\textwidth]{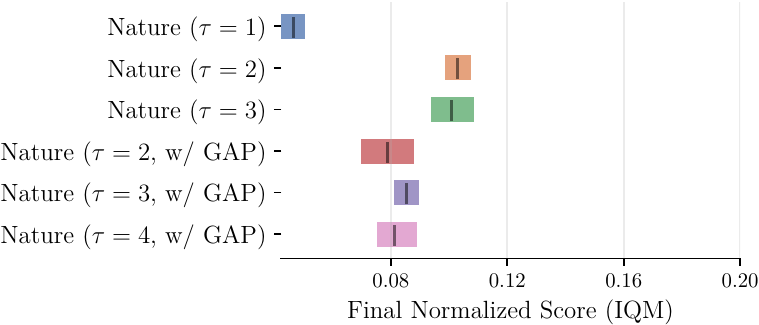}
        };
    \end{scope}
  \end{tikzpicture}
    \caption{Scaled Nature-CNN with results for generalization (\textit{easy}) using a subset of 10 Procgen games. The original Nature-CNN is shown as dotted lines; results with \gls*{gap} are solid. We show the final IQM scores (\textbf{bottom right}) and training curves with mean and standard deviation (\textbf{rest}).}
    \label{app:fig:nature_test}
\end{figure}

It can be seen in Figure~\ref{app:fig:nature_test} that for Nature-CNN, adding the \gls*{gap} layer is not beneficial. 
We hypothesize that the reason for this finding is that the Nature-CNN, even when increasing the network width, has too little depth and is still underparameterized.
The Nature-CNN with $\tau$=2 has 342,448 parameters in total, of which 262,400 are located in the Linear layer after flattening.
In contrast, when adding the \gls*{gap}, this Linear layer is reduced to 16,640 parameters, resulting in a total of merely 96,688 network parameters.
As such, the parameter reduction, which was beneficial for the overparametrized Impala-CNN, consisting of 15 layers and 1,441,680 parameters for $\tau$=2, may have a degrading effect, even if other network characteristics may be improved.

We plan further experiments with an altered, deeper network version of the Nature-CNN in future work, as a more complex experiment design is required to investigate this finding meaningfully.

\clearpage

\section{Additional Material for Experiments}
\label{sec:ap:experiments}
Learning curves based on the evaluation runs which run every 2.5M and 10M, respectively, time steps.
The results show the mean and standard deviation values of the normalized scores $S$ per Procgen environment, i.e., 1.0 corresponds to an optimal policy.

\begin{figure}[h]
    \centering
    \includegraphics[width=0.85\textwidth]{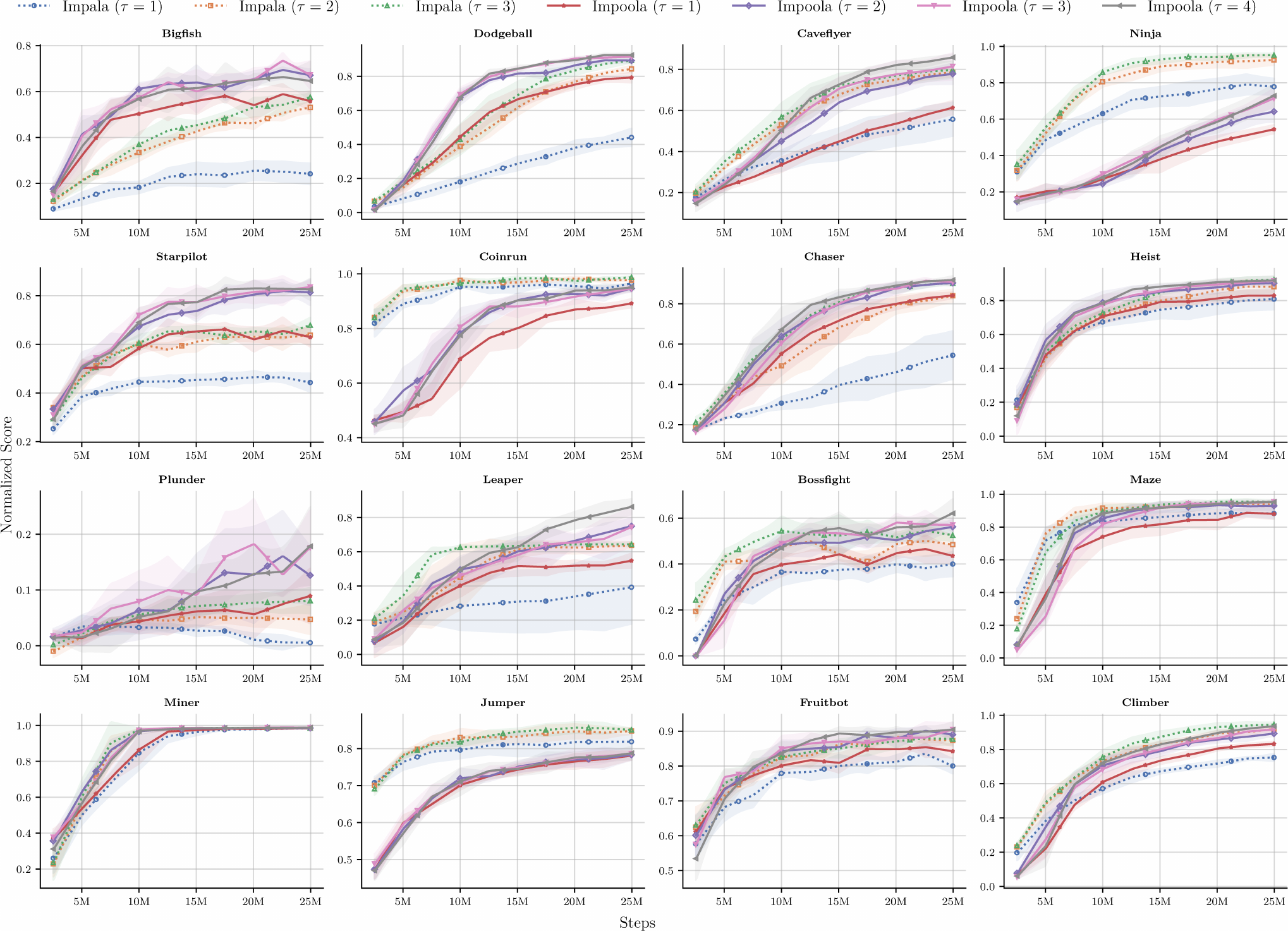}
    \caption{Detailed results for PPO with different width scalings $\tau$ in the \textit{easy} generalization track for \textit{training} levels. Impala-CNN is depicted as dotted lines.}
    \label{fig:detailed_results_ppo_scaling_training}
\end{figure}

\begin{figure}[h]
    \centering
    \includegraphics[width=0.85\textwidth]{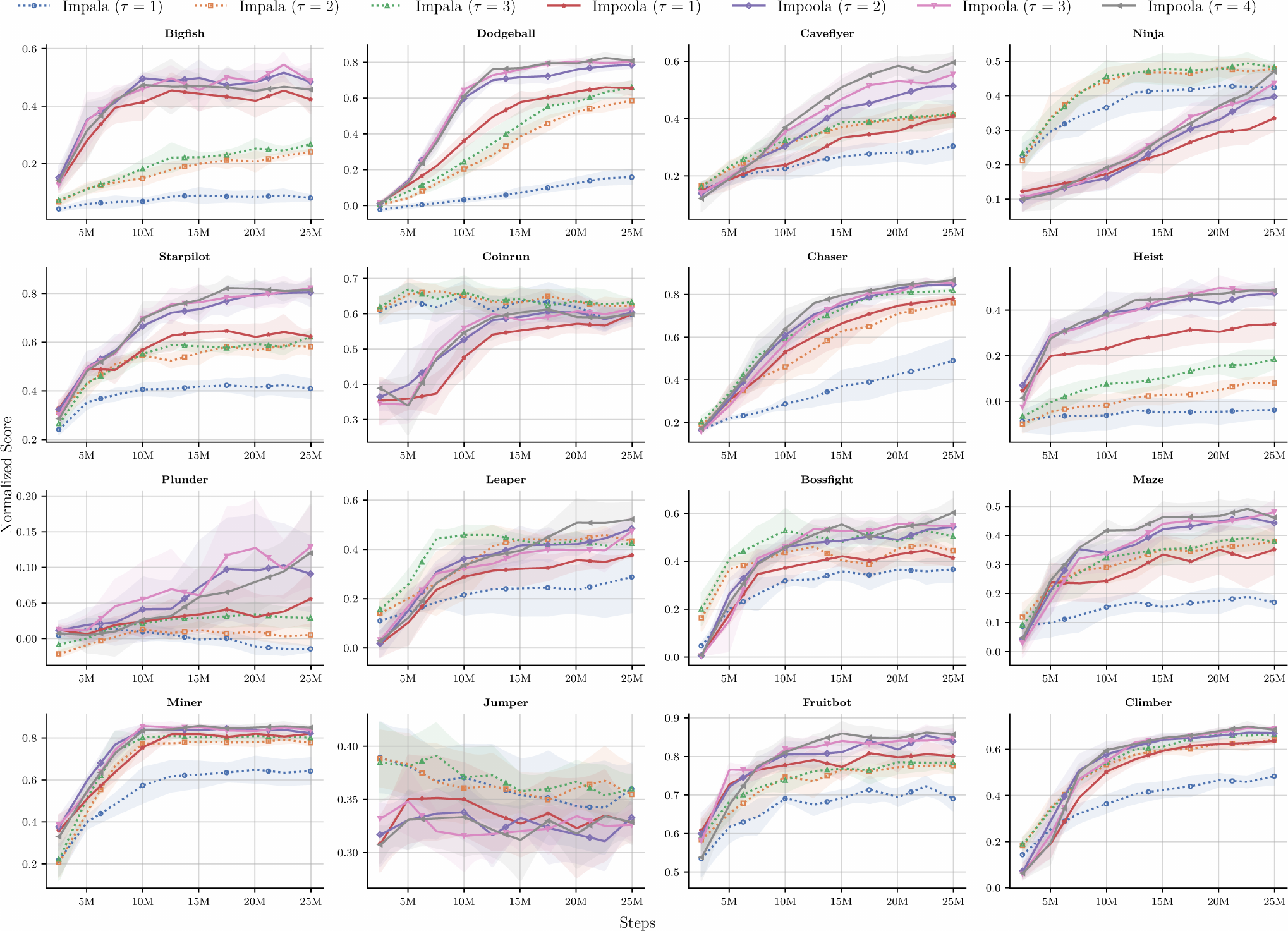}
    \caption{Detailed results for PPO with different width scalings $\tau$ in the \textit{easy} generalization track for \textit{testing} levels. Impala-CNN is depicted as dotted lines.}
    \label{fig:detailed_results_ppo_scaling_testing}
\end{figure}

\begin{figure}
    \centering
    \includegraphics[width=0.9\textwidth]{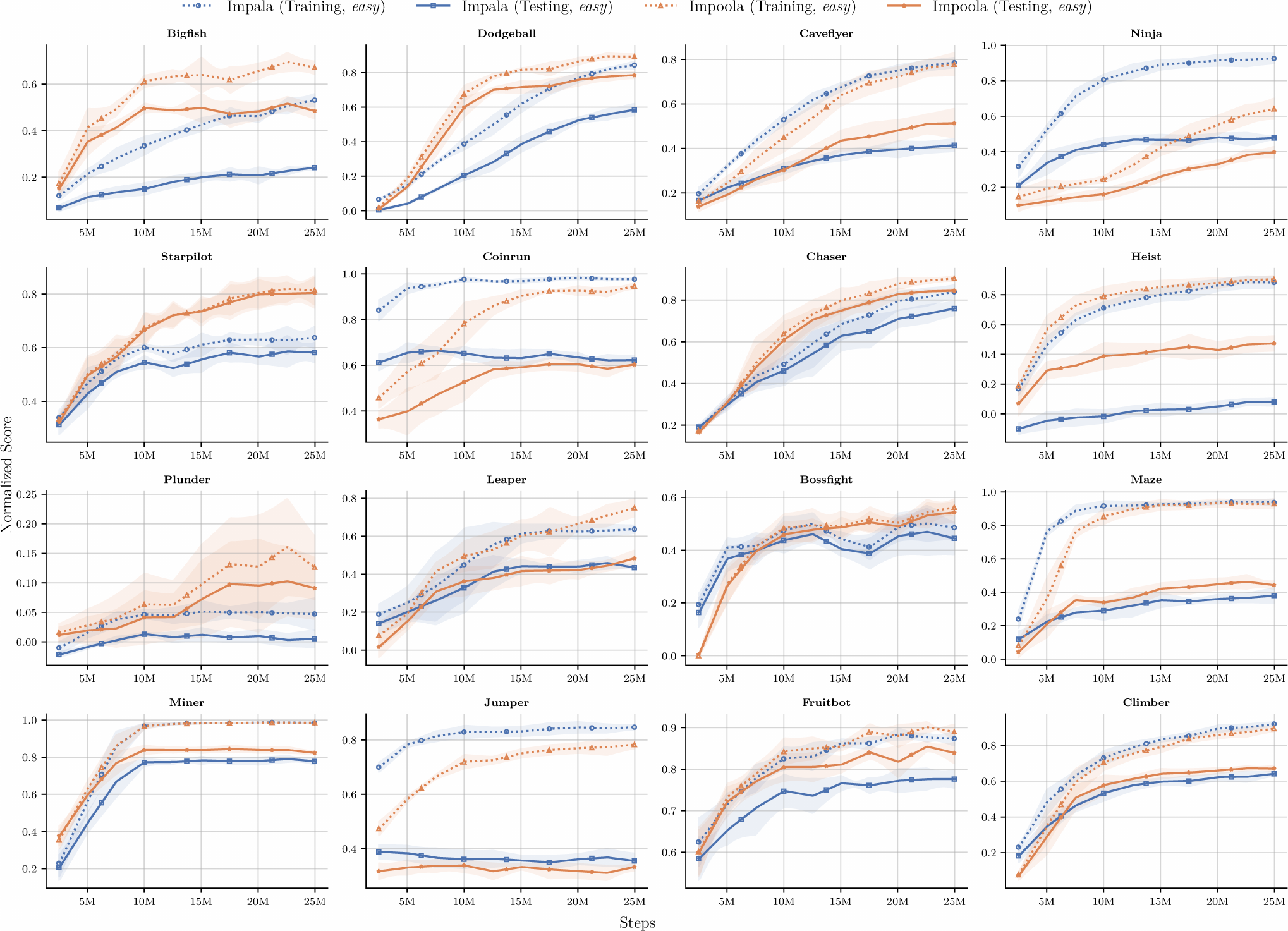}
    \caption{Detailed results for PPO ($\tau=2$) in the \textit{easy} generalization track for \textit{training} and \textit{testing} levels. Training levels are depicted as dotted lines.}
    \label{fig:detailed_results_ppo_generalization_easy}
\end{figure}

\begin{figure}
    \centering
    \includegraphics[width=0.9\textwidth]{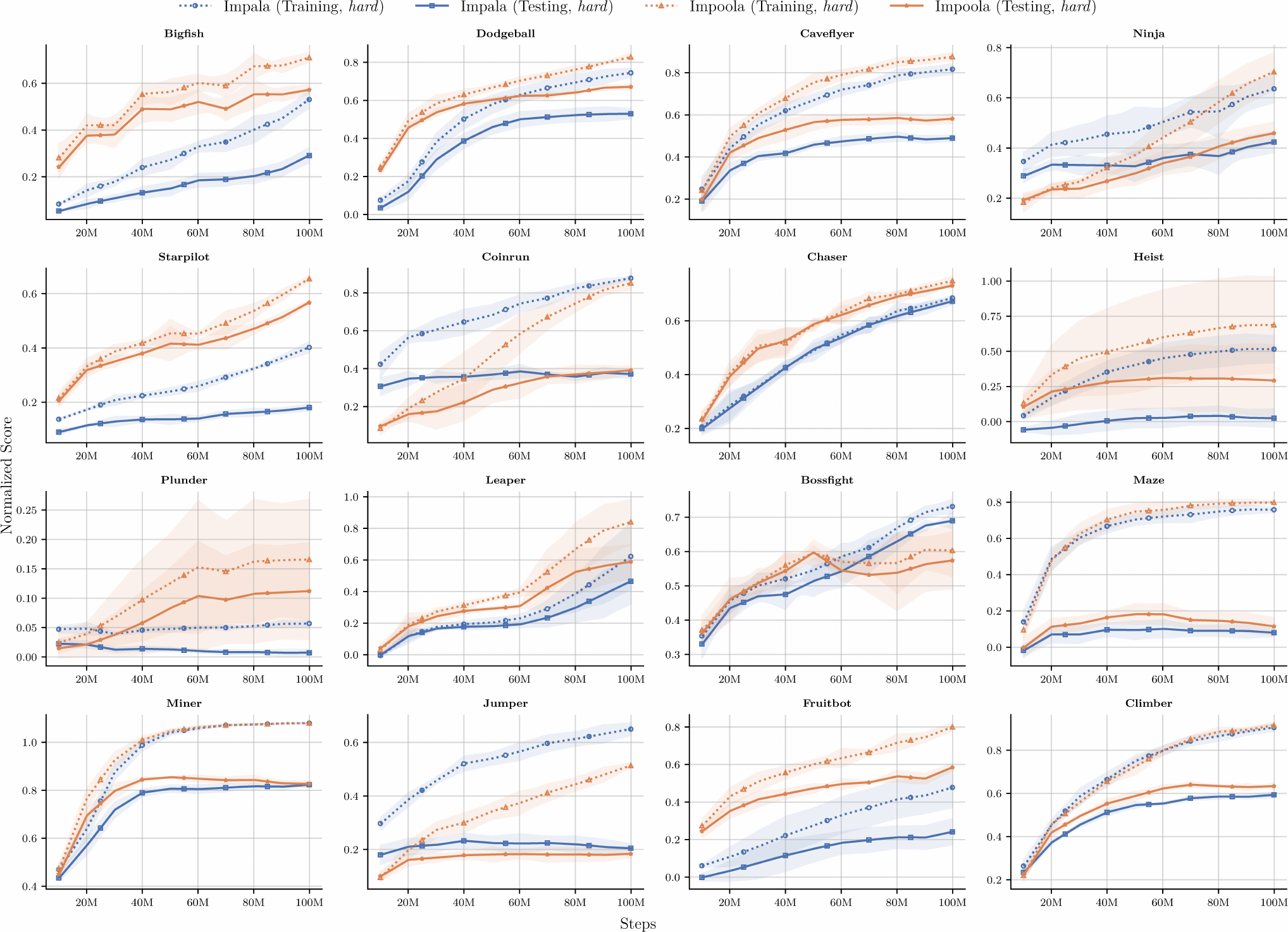}
    \caption{Detailed results for PPO ($\tau=2$) in the \textit{hard} generalization track for \textit{training} and \textit{testing} levels. Training levels are depicted as dotted lines.}
    \label{fig:detailed_results_ppo_generalization_hard}
\end{figure}

\begin{figure}
    \centering
    \includegraphics[width=0.9\textwidth]{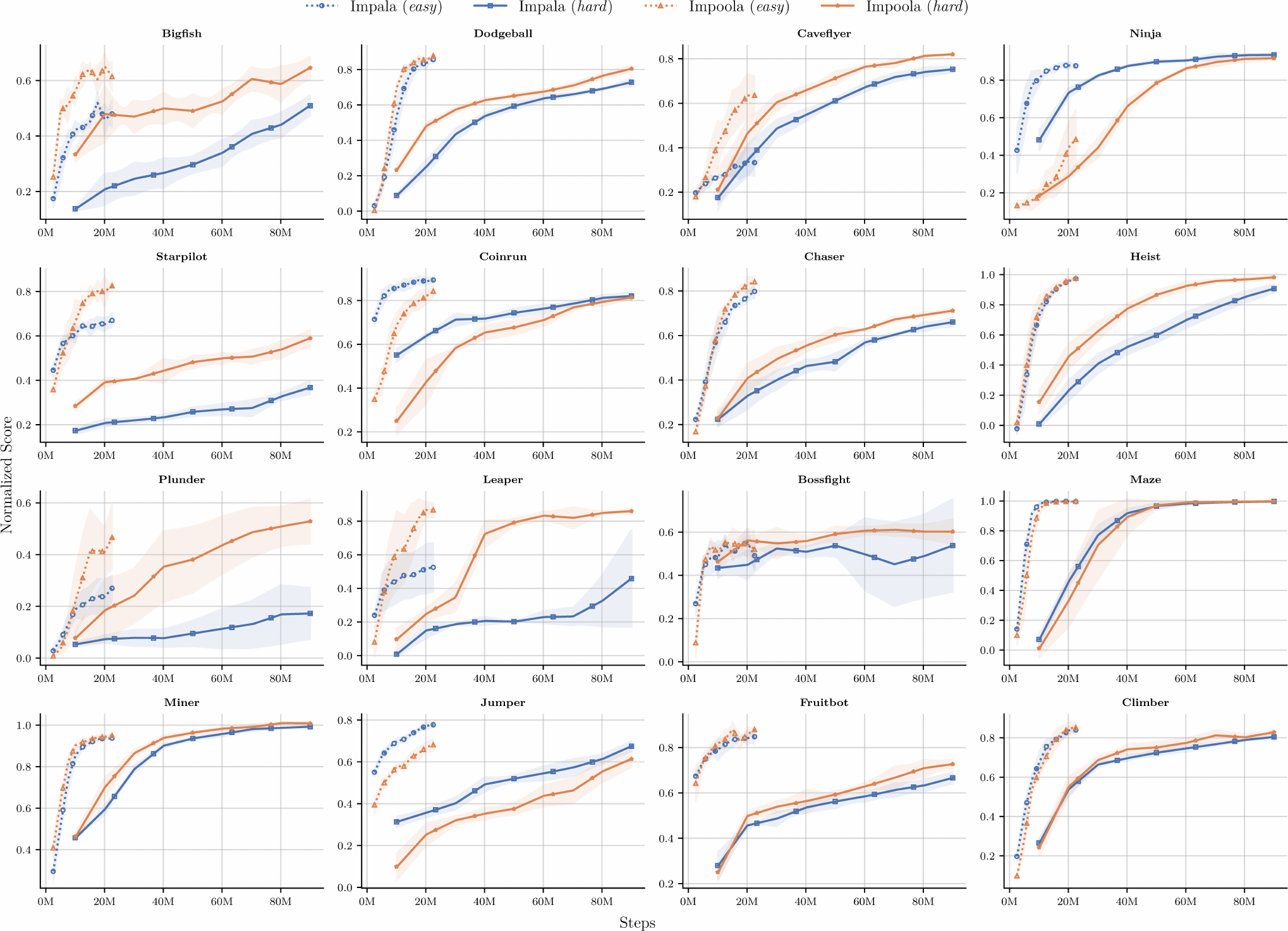}
    \caption{Detailed results for PPO ($\tau=2$) in the \textit{efficiency} track (\textit{easy} and \textit{hard}). Easy levels are depicted as dotted lines.}
    \label{fig:detailed_results_ppo_efficiency}
\end{figure}

\begin{figure}
    \centering
    \includegraphics[width=0.9\textwidth]{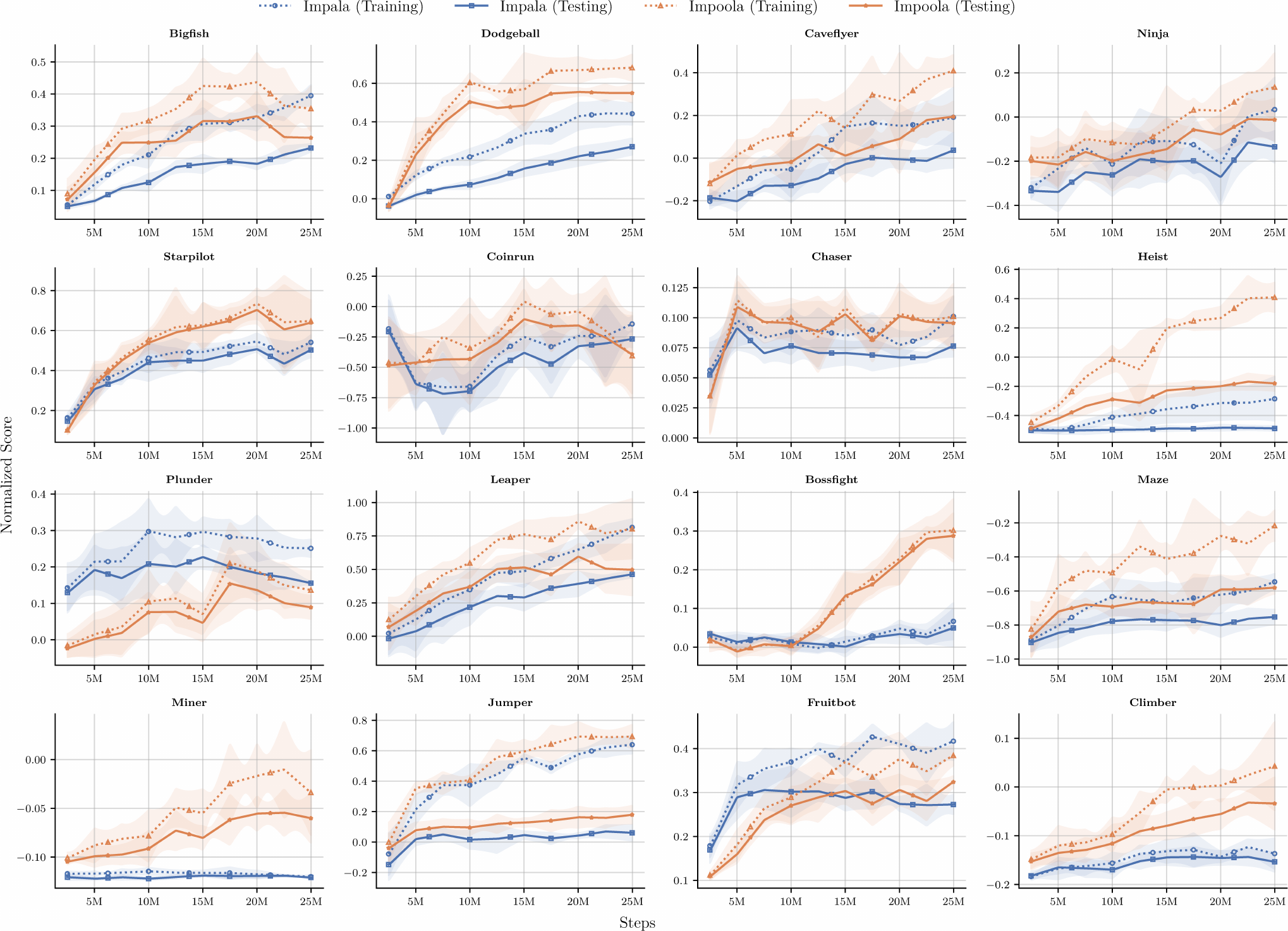}
    \caption{Detailed results for DQN ($\tau=2$) in the generalization track (\textit{easy}) for \textit{training} and \textit{testing} levels. Training levels are depicted as dotted lines.}
    \label{fig:detailed_results_dqn_generalization_easy}
\end{figure}

\begin{figure}
    \centering
    \includegraphics[width=0.9\textwidth]{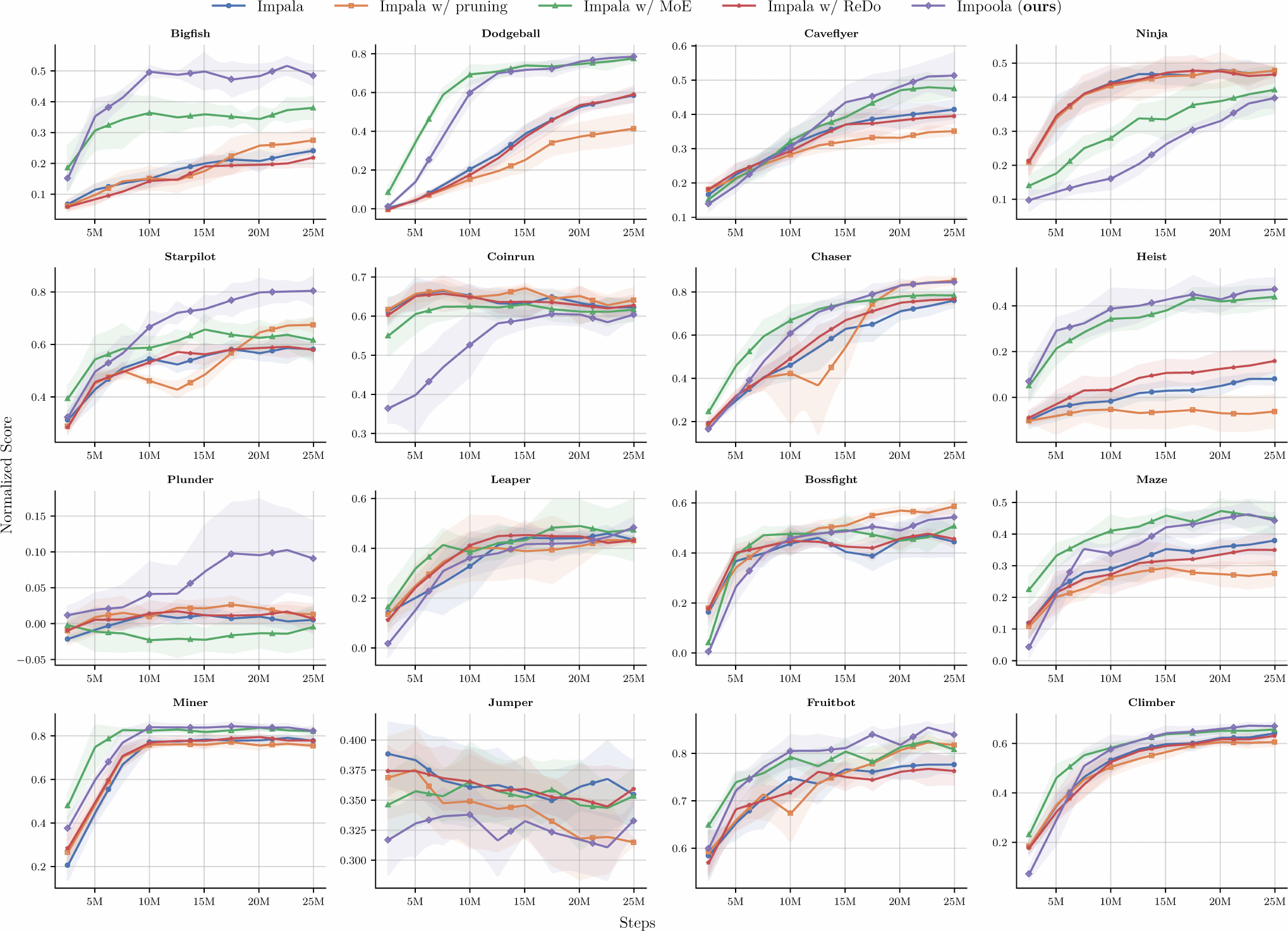}
    \caption{Detailed results for the benchmark of Impoola-CNN ($\tau=2$) against other methods in the  \textit{generalization} track (\textit{easy}) for \textit{testing} levels.}
    \label{fig:detailed_results_benchmark_generalization_easy}
\end{figure}


\end{document}